\documentclass[10pt,twocolumn,letterpaper]{article}
\pdfoutput=1
\usepackage{cvpr}              %
\usepackage{multirow}
\usepackage{amsmath}
\usepackage{amssymb}
\usepackage{pifont}
\usepackage{makecell}
\usepackage{graphics}
\usepackage{graphicx}

\usepackage[dvipsnames]{xcolor}

\definecolor{cvprblue}{rgb}{0.21,0.49,0.74}
\usepackage[pagebackref,breaklinks,colorlinks,citecolor=cvprblue]{hyperref}

\def\paperID{5211} %
\def\confName{CVPR}
\def\confYear{2024}

\title{Improve Supervised Representation Learning with Masked Image Modeling}

\author{Kaifeng Chen$^1$  \hspace{-0.4em}  \footnotemark[1] \hspace{-0.10em}
\and
Daniel Salz$^1$  \hspace{-0.4em} \footnotemark[1] \hspace{-0.10em}
\and
Huiwen Chang$^2$ \footnotemark[2]
\and
Kihyuk Sohn$^1$
\and
Dilip Krishnan$^1$
\and
Mojtaba Seyedhosseini$^1$ \\\\
\vspace{-5pt}
$^1$Google Research \hspace{30pt} $^2$OpenAI
}
\begin{document}
\maketitle
\renewcommand{\thefootnote}{\fnsymbol{footnote}}
\footnotetext[1]{Both authors contributed equally to this paper.}
\footnotetext[2]{Work done while at Google Research.}
\footnotetext{Correspondence: francischen@google.com, danielsalz@google.com}

\begin{abstract}
Training visual embeddings with labeled data supervision has been the de facto setup for representation learning in computer vision. Inspired by recent success of adopting masked image modeling (MIM) in self-supervised representation learning, we propose a simple yet effective setup that can easily integrate MIM into existing supervised training paradigms. In our design, in addition to the original classification task applied to a vision transformer image encoder, we add a shallow transformer-based decoder on top of the encoder and introduce an MIM task which tries to reconstruct image tokens based on masked image inputs. We show with minimal change in architecture and 
no overhead in inference that this setup is able to improve the quality of the learned representations for downstream tasks such as classification, image retrieval, and semantic segmentation. We conduct a comprehensive study and evaluation of our setup on public benchmarks. On ImageNet-1k, our ViT-B/14 model achieves $81.72\%$ validation accuracy, $2.01\%$ higher than the baseline model. On K-Nearest-Neighbor image retrieval evaluation with ImageNet-1k, the same model outperforms the baseline by $1.32\%$. We also show that this setup can be easily scaled to larger models and datasets. Code and checkpoints will be released.
\end{abstract}    
\section{Introduction}
\label{sec:intro}

Image representation learning is a core research area in computer vision. A learned mapping function (the image encoder), embeds image pixels into embedding vectors, or representations, so that similar images have high Euclidean similarities in the mapped embedding space. Representation learning powers a variety of applications such as face recognition \cite{liu2017sphereface,schroff2014facenet, 2018marginsoftmax, wang2018cosface}, fine-grained image retrieval and reranking \cite{wang2014finegrain,kihyuk2016,nikos2023uie, shao2023} and visual search at large scale \cite{2015pinterest, 2018bingsearch}. Modern Vision-Language Models (VLM) \cite{chen2023pali, wang2023beit3}, segmentation models \cite{kirillov2023sam} and conditional generative models \cite{shi2023instantbooth,li2023blipdiffusion, ramesh2022dalle2} also rely on well-trained image representations as inputs.

In a supervised learning setup, a dataset of images, each associated with a label, are usually collected through human raters, and a well-designed loss function supervises the deep learning model to predict these labels. One of the commonly used loss types is classification \cite{2018marginsoftmax, wang2018cosface, 2019arcface, zhai2019classification}, which enforces the representations to model the label distribution through a classifier layer. In computer vision, supervised representation learning has been successfully used on convolutional neural networks \cite{he2016resnet,NIPS2012_c399862d} and more recently, on vision transformers (ViT) \cite{alexander2021vit,zhao2022scalingvit}.
 
\begin{figure}[t]
\begin{center}
   \includegraphics[width=1.0\linewidth]{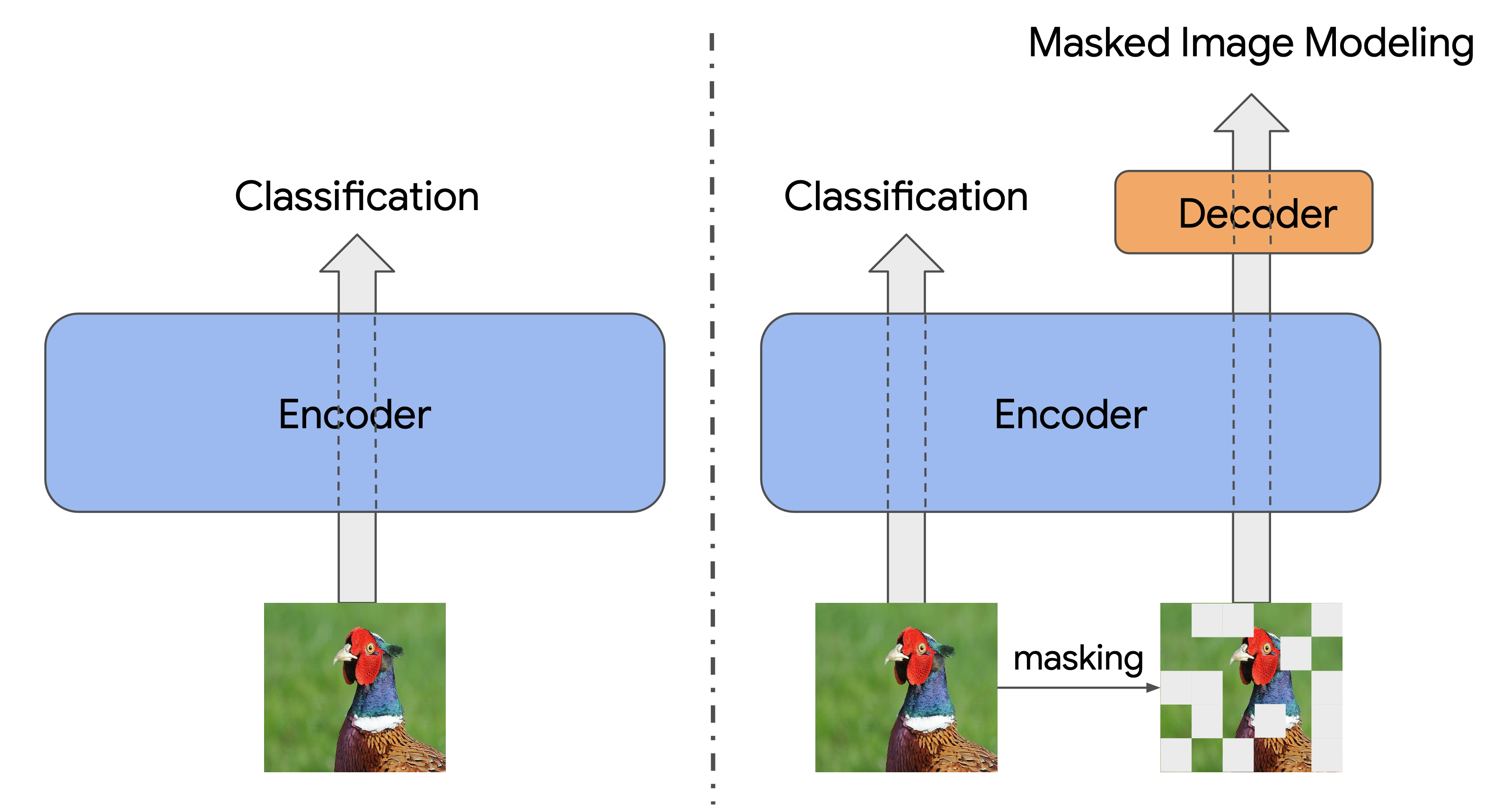}
\end{center}
\vspace{-10pt}
   \caption{Previous work (left) on representation learning has been mainly focused on using classification loss as the supervision. We propose a new setup (right) that combines supervised learning with masked image modeling. During training, in addition to standard classification supervision, the input images are masked and encoded by the same image encoder, and a shallow decoder is trained to decode the encoded patches %
   back to low-level image tokens or pixels %
   . At inference time, the decoder is thrown away, and images are classified via the encoder.
   \vspace{-15pt}
   }
\label{fig:fig1}
\end{figure}

In addition to supervised representation learning that heavily relies on image labels, self-supervised representation learning \cite{2020simclr, Caron2021dino, bao2022beit, kaiming2022CVPR}, which works without necessitating any labels in pre-training, has gained popularity with recent successes. Among different self-supervised learning paradigms, masked image modeling (MIM) \cite{bao2022beit, kaiming2022CVPR, Li2023mage} has attracted significant research interests given its resemblance to language model pre-training \cite{2019bert}, where input images are masked out according to a carefully optimized strategy and a deep learning model is trained to reconstruct masked pixels or image tokens. Similar to 
language models, the task of learning to predict missing tokens is a powerful inductive bias that leads to excellent image representations for downstream tasks.

Supervised training uses a single semantic label for each image; MIM, on the other hand, tries to predict a large number of low-level token-level details. Intuitively, these two losses seem to be complementary to each other. This observation inspires us to think about the following question: \textbf{Can we incorporate MIM in supervised representation learning and further improve the learned image representations with no significant training or inference overheads?} To answer this, we propose a new setup on the right of Figure~\ref{fig:fig1}. In addition to the traditional classification-only supervised representation learning diagram (left in Figure~\ref{fig:fig1}), we introduce an MIM task that masks image inputs, encodes them to produce masked representations, and decodes image tokens using the representations. Given the original image encoder is shared between the supervised learning task and the MIM task, the decoder is shallow, and no extra training data is introduced, this setup thus imposes minimal change to any existing architecture. During test time, since only the trained encoder is needed to compute the representations, our setup has the same inference cost as standard supervised models trained without MIM.

In this paper, we mainly focus our discussions on the ViT model family. We provide detailed description of our setup in Section~\ref{sec:method} and studies on standard size ViT models and their performance on classification, image retrieval,  and semantic segmentation evaluations in Section~\ref{sec:results}, and include the transfer learning \cite{zhai2020vtab} in the supplementary material. Our main contributions are as below.

\noindent\textbf{Contributions.} (1) We propose a simple and effective training setup that combines supervised representation learning with MIM. Our setup consists of an image encoder and a shallow decoder, each image is processed via two tasks: a standard supervised learning task with classification loss, and an MIM task that encodes the masked image and decodes image tokens. The aggregation of the two losses from both tasks is applied to the network to guide the training. (2) We show that the learned image representation is significantly improved compared with the baseline model without the MIM task. In particular, our ViT-B/14 model trained on ImageNet-1k \cite{degn2009imagenet} outperforms the baseline model by $2.01\%$ in ImageNet-1k validation accuracy, $1.32\%$ on ImageNet retrieval task and $1.21\%$ in semantic segmentation on the ADE20k \cite{zhou2017ade20k} dataset. %
(3) We provide comprehensive ablation study on the choice of parameters for training, and show our approach can be easily extended to larger models and datasets. 
\section{Related Work}
\label{sec:rw}

\noindent
\textbf{Supervised Representation Learning in Computer Vision.}
Using provided dataset labels as supervision, this approach aims to learn embeddings that preserve the similarity between samples from the same class. Typical methods to train such an embedding space is to apply metric learning losses such as contrastive \cite{lecun2005}, triplet \cite{tripletloss, schroff2014facenet} and classification \cite{wang2018cosface, 2019arcface, NIPS2012_c399862d, zhai2019classification}. These losses either minimize distances between similar images and maximize distances between dissimilar images, or penalize wrong class predictions. The choice of loss also depends on the type of labels. Among these losses, classification is widely adopted in face recognition \cite{schroff2014facenet}, image retrieval \cite{kihyuk2016} and general representation learning \cite{zhai2019classification}, and achieves state-of-the-art performance. Moreover, latest ViT \cite{alexander2021vit} models mainly use classification loss for their training and evaluation benchmarks.

\noindent
\textbf{Self-supervised Representation Learning and Masked Image Modeling.} 
Acquiring labeled datasets is expensive. An alternative approach that does away with labels, is to use self-supervised representation learning. It has proven to be viable and effective to learn image representations only using image pixels, resulting in competitive performance to that of supervised pre-training. One family of discriminative self-supervised learning is contrastive learning, where a model is trained based on pairs of ``positive" and ``negative" images, and several methods are proposed to improve the sampling of negatives. For instance, SimCLR \cite{2020simclr} proposes to mine negative samples within each batch when combined with large batch sizes; momentum-contrastive approach (MoCo) \cite{he2020moco} introduces a novel training method to support a large number of negative samples by leveraging a separate encoder with moving average weights and a queue for storage. Other methods propose different approaches to build contrastive pairs. BYOL \cite{grill2020boyl}, for example, relies on two neural networks to learn from each other during training. SwAV \cite{mathilde2020swav} predicts cluster assignment of a view by using the representations of different views. More recently, DINO \cite{Caron2021dino} shows that using a momentum teacher and proper augmentation yields better scaling property, followed by several improvements made in DINOv2 \cite{oquab2023dinov2}.

Aside from contrastive loss, an alternative approach that has gained more popularity in research recently is masked image modeling (MIM). Inspired by masked language modeling of BERT \cite{2019bert}, MIM uses masked images as the input and learns to reconstruct the information of the original image. MIM relies on intra-image relationships and naturally works with transformer \cite{Vaswania2017attention} architectures.  Several works show success along this direction. In BEiT's design \cite{bao2022beit, peng2022beit}, the reconstruction targets are visual tokens learned through a separate image quantizer also trained in a self-supervised way. PeCo \cite{peco2023} improves upon BEiT by adopting a better tokenizer. MAGE \cite{Li2023mage} combines token reconstruction with contrastive loss to enable both representation learning and image generation. Mased Autoencoder (MAE) \cite{kaiming2022CVPR} employs an encoder-decoder architecture and uses pixel-level reconstruction as the task. Image features generated by teacher models are also considered as reconstruction targets in MaskFeat \cite{wei2022maskfeat} and MVP \cite{wei2022mvp}.

Self-supervised approaches are usually used for pre-training, which follows a fully supervised fine-tuning stage to reach the best performance. Different from this training process, in our design, we propose to train them together in one stage which yields better performance than supervised training alone.

\noindent
\textbf{Training ViT Models on ImageNet.}
Training better models on ImageNet dataset \cite{degn2009imagenet} has been one of the long standing challenges in computer vision. Extensive explorations have been conducted on recent ViT models. When only having classification, AugReg \cite{steiner2022augreg} provides a comprehensive study on the impact of model regularization and data augmentation on training ViT models. DEiT \cite{deit} shows significant improvements in ImageNet-1k validation accuracy when tasked with a distillation target. However, the training losses for these works are either direct supervision or using soft labels produced from supervision. %
In the scope of this work, we build upon the prior art and explore adding unsupervised tasks for further improvements.

\section{Method}
\label{sec:method}

We propose a training framework composed of two objectives: supervised classification objective (i.e., cross-entropy loss) and self-supervised MIM objective. Two streams share an encoder, while a decoder is introduced to predict masked patches or tokens for MIM objective, which we explain in Section~\ref{subsec:encoder}.
Section~\ref{subsec:sltask} and \ref{subsec:mim} introduces learning tasks and Section~\ref{subsec:training} explains our final training objective.

\begin{figure*}[t]
\begin{center}
   \includegraphics[width=0.95\linewidth]{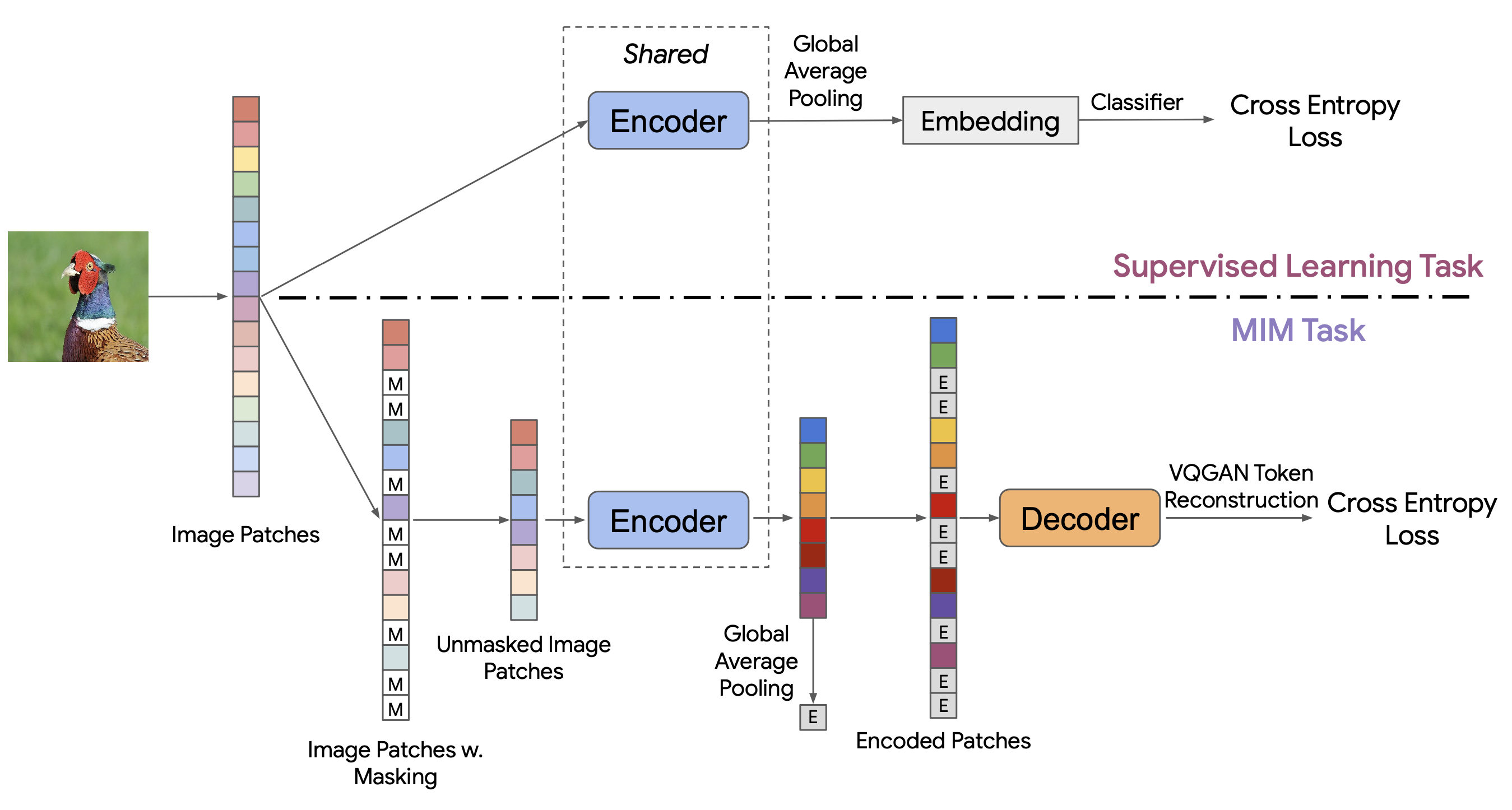}
\end{center}
\vspace{-5pt}
   \caption{Details of our setup. For the supervised learning task, we follow the standard convention of training ViT models with cross entropy loss (with no masking on input patches). For the MIM task, masked images are encoded with the shared ViT encoder and decoded to predict image tokens.
   }
\label{fig:fig2}
\end{figure*}

\subsection{Encoder and Decoder Design}
\label{subsec:encoder}
ViT \cite{alexander2021vit} backbone is used as a shared image encoder. Each input image is reshaped into a sequence of flattened 2D patches or tokens, with the shape of each being $p\times p \times C$, where $C$ is the number of channels, and then the sequence is projected to patch embeddings. We use standard learnable 1D position embeddings to encode positional information into the patch embeddings \cite{alexander2021vit}, followed by several transformer layers to encode the patches. Instead of using the output of a learnable \texttt{[CLS]} token as the image representation as done in \citep{alexander2021vit}, we follow \cite{zhao2022scalingvit} and performed global average pooling (GAP), which averages the encoded token embeddings at the transformer output to produce a vector representation per image. We consider ViT-B and ViT-L size models, which are considered reasonable sizes for real world deployment.

The decoder is based on the transformer architecture. The input sequence length of the decoder is identical to that of the encoder to enable training the MIM task (see Section \ref{subsec:mim}). Pre-LayerNorm is used inside the transformer blocks following the original transformer implementation \cite{Vaswania2017attention}. Moreover, we use the same set of parameters as the ViT-B model except we use only 1 transformer layer. Compared to larger architectures, the choice of a light-weight decoder reduces training overhead and retains performance. We provide an ablative study in Section~\ref{subsec:ablation}. Details of the encoder and decoder parameters are discussed in Section \ref{sec:results}.

\subsection{Supervised Learning Task}
\label{subsec:sltask}
The upper half of Figure ~\ref{fig:fig2} describes the setup of the supervised learning task. Following \cite{wightman2021resnet,hugo2022,zhao2022scalingvit}, we formulate the training objective as a multi-class classification problem and use sigmoid cross entropy loss for the classification task. Assuming an image $x_i$ is associated with label $y_i$ and logits $l_i$, the loss is formulated as:
\begin{equation}
\label{eq1}
    L_{\text{CE}}(x_i)=-y_i\log(\sigma(l_i)) - (1-y_i)\log(1-\sigma(l_i)),
\end{equation}
where $\sigma(x) = 1/(1+e^{-x})$ is the sigmoid function. %

\subsection{MIM Task}
\label{subsec:mim}
Our MIM task requires masked images at the encoder inputs and image tokens as the reconstruction target. The details of each component are discussed in detail in this section.

\textbf{Masking Ratio.}
In most MIM works, dynamic masking ratio strategy has proven to be effective. BEiT \cite{bao2022beit} adopts a block-wise masking algorithm with masking ratio of at least $40\%$. More sophisticated masking ratio is obtained in MAGE \cite{Li2023mage} through the combination of truncated Gaussian distribution and random masking. In our work, however, we find our models trained with a constant masking ratio are already performant. We attribute this to the unique setting of co-training supervised learning and MIM.
 
\textbf{VQGAN Tokenizer.} 
Image tokens are integer latents that can represent image pixels through an encoder and decode original pixels through a decoder. Also trained in a self-supervised way, the tokens encode high-level semantic information without losing low-level details, thus being suitable for generative models \cite{chang2023muse, yu2022vectorquantized, Chang2022maskgit}. Recent generative pre-training works \cite{bao2022beit, peng2022beit, Li2023mage} also use image tokens as the prediction target and show superior results in representation learning. We follow the same idea here and use the encoder part of a VQGAN \cite{esser2021vqgan} that was trained on ImageNet-1k to produce image tokens, which are used as the prediction target of our decoder. Our setup requires the number of latent tokens produced by the VQGAN to match the sequence length of the encoder and decoder to keep masking and prediction consistent with the tokens. %
 
\textbf{Reconstructive Training.}
As is shown in the lower half of Figure \ref{fig:fig2}, after applying masking to the input sequence (mask tokens are denoted by $\texttt{[M]}$), only the unmasked image patches are passed to the encoder. At the output of the encoder we produce a representation, denoted by $\texttt{[E]}$, by applying GAP to the unmasked encoded patches. Similar to \cite{Li2023mage}, the encoded patches are then re-arranged according to its original position, where the original masked positions are replaced by $\texttt{[E]}$. For an image $x_i$ and a sequence length $N$, we denote VQGAN tokens $T_i=[t_i^j]|_{j=1}^N$ and binary mask $M_i=[m_i^j]|_{j=1}^N$ that determines which tokens to mask out. The training objective is then to reconstruct tokens from the unmasked inputs %
. The cross entropy loss between the ground-truth one-hot tokens and the output of the decoder is specified as
\begin{equation}
\label{eq2}
    L_{\text{MIM}}(x_i)=-\sum_{j} \log p(t_i^j|T_{M_i} ),
\end{equation}
where $T_{M_i}$ are the unmasked tokens in $T_i$ and $p(t_i^j|T_{M_i} )$ is the predicted probability conditioned on the unmasked tokens. Unlike the observations in \cite{kaiming2022CVPR, Li2023mage}, we found applying loss on \textbf{all} tokens works better than only on masked tokens in our case%
. Ablations on only predicting masked tokens in the loss are provided in the supplementary material.

\subsection{Training Procedure}
We train supervised learning and MIM tasks together in a single stage, with the following form for the total loss
\begin{equation}
\label{eq3}
    L(x_i)=L_{\text{CE}}(x_i)+ \lambda\cdot L_{\text{MIM}}(x_i),
\end{equation}
where $\lambda$ is the weight on the MIM task. When $\lambda=0$, the above training objective will result in the baseline model supervised with only classification loss.
\label{subsec:training}

During training, we keep using the \textbf{same} batch for both tasks. That is, after sampling one batch, the same set of images will be used to calculate the losses defined in Equations \ref{eq1} and \ref{eq2}. As a result, training for each batch consists of two rounds of forward propagation for both tasks and one round of backward propagation. This greatly simplifies the implementation of sampling independent batches for each task and makes it straightforward to be plugged into existing image encoders. Furthermore, the same learning rate schedule and optimizer is applied to both tasks, leaving $\lambda$ the only hyperparameter that balances the weights between the two types of loss.

At test time, only the encoder is used for inference: unmasked images are the inputs to the encoder and the output after applying GAP will be the image representation.

\section{Results}
\label{sec:results}
We describe the training details and the evaluation results of our method. We use a JAX/Flax \cite{dehghani2021scenic} codebase for pre-training and fine-tuning tasks using v3 TPUs. All datasets are accessed through Tensorflow Datasets library \cite{tensorflow2015-whitepaper} to ensure reproducibility. We mainly focus on training ViT-B and ViT-L models on ImageNet-1k and ImageNet-21k datasets to better understand the scaling property of our approach in terms of model size and data size. As our goal is to measure the quality of the learned representation, we evaluate our models on commonly used benchmarks such as classification, retrieval and several downstream evaluations.

\subsection{Training Details}
\label{subsec:eval}
Following \cite{steiner2022augreg}, we adopt the same data augmentation and model regularization for both the baseline and our models. Concretely, all training experiments are done on $224\times 224$ resolution images with patch size $p=14$, leading to input sequence length of $256$. We take the off-the-shelf VQGAN \cite{esser2021vqgan} model trained on ImageNet-1k with vocabulary size $1024$, which takes in image resolution of $256\times 256$ and output $16\times 16$ tokens, matching the sequence length of the encoder and decoder. The tokens of the images are computed on the fly after the augmentation. Similar to \cite{kaiming2022CVPR}, we find using a \textbf{single} transformer layer decoder already provides strong results. Through comprehensive ablation, we also find $20\%$ masking gives the best trade-off between quality and efficiency. The detailed ablation studies of the number of transformer layers in the decoder and masking ratio are provided in Section \ref{subsec:ablation}. We use $\lambda=1$ in Equation \ref{eq3} in all our experiments, i.e., equally weighting the supervised and MIM losses, to avoid heavy hyperparameter sweep and keep the training recipe simple. Other pre-training hyperparameters are specified in Table \ref{table:training_param}. The baseline models for all the following comparisons are trained only on the classification task, but otherwise use the same the training settings as our models.

\begin{table}[h]
\centering
\begin{tabular}{c|c|c}
\toprule
Parameter & ViT-B/14  & ViT-L/14\\
\midrule
Batch size & \multicolumn{2}{c}{4096} \\
Optimizer & \multicolumn{2}{c}{AdamW \cite{loshchilov2018adamw}} \\
Learning rate & \multicolumn{2}{c}{1e-3} \\
$\beta_1$/$\beta_2$ & \multicolumn{2}{c}{0.9/0.96} \\ 
ImageNet-1k & \multicolumn{2}{c}{w. AugReg \cite{steiner2022augreg}}\\
ImageNet-21k & \multicolumn{2}{c}{w/o. AugReg \cite{steiner2022augreg}}\\
Learning rate warm up/decay & \multicolumn{2}{c}{10000 steps/cosine}  \\
Training epochs & \multicolumn{2}{c}{30/300}\\
\midrule
Weight decay & 0.03 & 0.1 \\
\bottomrule
\end{tabular}
\vspace{-5pt}
\caption{Pre-training parameters for ViT-B and ViT-L models.}
\label{table:training_param}
\end{table}

We use the notation \texttt{IN-1k$_{300}$} for ImageNet-1k pre-training for 300 epochs. \texttt{IN-21k$_{30}$} and \texttt{IN-21k$_{300}$} are for models pre-trained on ImageNet-21k for 30 and 300 epochs, respectively. \texttt{IN-21k$_{30}$ FT} and \texttt{IN-21k$_{300}$ FT} represent \emph{fine-tuning} the corresponding pre-trained models with ImageNet-1k dataset. For ImageNet-1k fine-tuning, we train the model for 20000 steps with batch size $512$ and learning rate 3e-2 without AugReg.  Note for ViT-B/14 models, we consider both ImageNet-1k and ImageNet-21k pre-training.  While for ViT-L/14 models, following common practice, we only do ImageNet-21k pre-training.

\subsection{ImageNet Classification}
\label{subsec:classification}
For ImageNet-1k classification we follow standard practice and measure accuracy on the validation set. We present the results in Table \ref{table:classification_eval}.  

\begin{table}[h]
\centering
\scalebox{0.77}{
\begin{tabular}{l|ccc|ccc} 
\toprule
 \multirow{2}*{Training Setup} & \multicolumn{3}{c|}{ViT-B/14} & \multicolumn{3}{c}{ViT-L/14} \\
 \cmidrule{2-7}
 & Baseline & Ours & $\Delta$ & Baseline  & Ours& $\Delta$\\
 \midrule
IN-1k$_{300}$ & 79.71 & 81.72 & \textbf{+2.01} & -- & -- & -- \\
IN-21k$_{30}$ FT & 81.50 & 82.46 & \textbf{+0.96} & 83.35 & 84.51 & \textbf{+1.16}  \\
IN-21k$_{300}$ FT & 82.82 & 83.68 & \textbf{+0.86} & 82.29 & 83.57 & \textbf{+1.28}  \\
\bottomrule 
\end{tabular}
}
\vspace{-5pt}
\caption{ImageNet-1k validation accuracy for various ViT-B/14 and ViT-L/14 models. Values are in percentage. We see a consistent boost over a well-tuned baseline, across architectures and training data distributions.}
\label{table:classification_eval}
\end{table}
We observe a consistent gain from our models. In particular, the accuracy is improved by $2.01\%$ with ViT-B/14 model trained on ImageNet-1k. Note this gain can be even higher when using $5\%$ masking ratio and a decoder with two transformer layers, as reported in the ablation study section \ref{subsec:ablation}. On larger models such as ViT-L, the gain is as significant as $1.28\%$ after pre-training and fine-tuning, showing our method is effective in scaling to larger models and datasets. Similar trends on ImageNet-Real \cite{imagenet_real} and ImageNet-v2 \cite{imagenet_v2} are observed as shown in Table \ref{table:classification_eval_more}, indicating that adding MIM task also greatly improves robustness of the learned representation.
\begin{table}[h]
\centering
\scalebox{0.85}{
\begin{tabular}{c|c|c|c|c} 
\toprule
\multicolumn{2}{c|}{Model} & Training Setup &  ImageNet-Real & ImageNet-v2 \\
\midrule
\multirow{9}*{\rotatebox[origin=c]{90}{ViT-B/14}}  & Baseline & \multirow{3}*{IN-1k$_{300}$} & 78.82 & 65.86 \\
  & Ours & & 80.77 & 68.52\\
 & $\Delta$  & & \textbf{+1.95} & \textbf{+2.66} \\
 \cmidrule{2-5}
 & Baseline & \multirow{3}*{IN-21k$_{30}$FT} & 86.23 & 68.97 \\
  & Ours & & 87.37 & 71.16 \\
 & $\Delta$  & & \textbf{+1.14} & \textbf{+2.19} \\
\cmidrule{2-5}
  & Baseline & \multirow{3}*{IN-21k$_{300}$ FT} & 86.61 & 70.83 \\
 & Ours &  & 87.61 & 72.15 \\
 & $\Delta$  & & \textbf{+1.00} & \textbf{+1.32} \\
\midrule
\multirow{6}*{\rotatebox[origin=c]{90}{ViT-L/14}}  & Baseline & \multirow{3}*{IN-21k$_{30}$FT} & 87.13 & 71.45 \\
  & Ours & & 88.39 & 73.63 \\
 & $\Delta$  & & \textbf{+1.26} & \textbf{+2.18} \\
\cmidrule{2-5}
  & Baseline & \multirow{3}*{IN-21k$_{300}$ FT} & 85.67 & 69.84 \\
 & Ours &  & 87.11 & 71.99 \\
 & $\Delta$  & & \textbf{+1.44} & \textbf{+2.15} \\
\bottomrule 
\end{tabular}
}
\vspace{-5pt}
\caption{ImageNet-Real validation and ImageNet-v2 test accuracy for the baseline and our ViT-B/14 and ViT-L/14 models. Numbers are in percentage. The consistent gains show the increase in robustness achieved by adding an MIM loss.}
\label{table:classification_eval_more}
\end{table}

\subsection{Image Retrieval}
\label{subsec:retrieval}

\begin{table*}[h]
\centering
\scalebox{0.75}{
\begin{tabular}{c|c|c|c|ccccccccc} 
\toprule
\multicolumn{2}{c|}{\multirow{2}*{Model}} & \multirow{2}*{Training Setup} & \multirow{2}*{ImageNet} & \multicolumn{9}{c}{Universal Embedding Dataset} \\
\cmidrule{5-13}
\multicolumn{2}{c|}{}  & & & Food2k & CARS196 & SOP & Inshop & iNat & Met & GLDV2 & Rp2k & Mean \\
\midrule
\multirow{9}*{\rotatebox[origin=c]{90}{ViT-B/14}} & Baseline& \multirow{3}*{IN-1k$_{300}$} & 78.15 & 26.31 & 45.34 & 63.62 & 46.05 & 47.38 & 20.34 & 14.17 & 72.42 & 41.95 \\
 & Ours  & & 79.47 & 28.27 & 51.43 & 62.40 & 43.76 & 52.90 & 26.92 & 17.18 & 71.78 & 44.33 \\
 & $\Delta$  & & \textbf{+1.32} & \textbf{+1.96} & \textbf{+6.09} & -1.22 & -2.29 & \textbf{+5.52} & \textbf{+6.58} & \textbf{+3.01} & -0.64 & \textbf{+2.38} \\
\cmidrule{2-13}
 & Baseline& \multirow{3}*{IN-21k$_{30}$ FT} & 78.08 & 33.71 & 47.77 & 51.60 & 37.70 & 63.41 & 16.75 & 15.94 & 59.77 & 40.83 \\
 & Ours  & & 79.14 & 36.61 & 52.32 & 53.20 & 40.15 & 68.20 & 20.23 & 17.80 & 60.73 & 43.66 \\
 & $\Delta$  & & \textbf{+1.06} & \textbf{+2.90} & \textbf{+4.55} & \textbf{+1.60} & \textbf{+2.45} & \textbf{+4.79} & \textbf{+3.48} & \textbf{+1.86} & \textbf{+0.96} & \textbf{+2.82} \\
\cmidrule{2-13}
 & Baseline& \multirow{3}*{IN-21k$_{300}$ FT} & 80.31 & 28.64 & 46.87 & 49.37 & 38.88 & 61.19 & 10.87 & 15.59 & 55.71 & 38.39 \\
 & Ours  & & 81.12 & 29.51 & 51.56 & 50.60 & 39.94 & 64.44 & 14.46 & 16.49 & 59.18 & 40.76 \\
 & $\Delta$  & & \textbf{+0.81} & \textbf{+0.87} & \textbf{+4.69} & \textbf{+1.23} & \textbf{+1.06} & \textbf{+3.25} & \textbf{+3.59} & \textbf{+0.80} & \textbf{+3.47} & \textbf{+2.37} \\
\midrule
\multirow{6}*{\rotatebox[origin=c]{90}{ViT-L/14}}  & Baseline & \multirow{3}*{IN-21k$_{30}$ FT} & 80.80 & 32.42 & 47.76 & 54.00 & 44.27 & 63.78 & 15.95 & 16.38 & 63.88 & 42.31 \\
  & Ours & & 82.06 & 35.03 & 53.57 & 54.46 & 45.98 & 68.38 & 21.78 & 20.64 & 63.17 & 45.38 \\
 & $\Delta$  & & \textbf{+1.26} & \textbf{+2.61} & \textbf{+5.81} & \textbf{+0.46} & \textbf{+1.71} & \textbf{+4.60} & \textbf{+5.83} & \textbf{+4.26} & -0.71 & \textbf{+3.07} \\
\cmidrule{2-13}
  & Baseline & \multirow{3}*{IN-21k$_{300}$ FT} & 81.16 & 22.94 & 33.65 & 46.65 & 42.76 & 55.14 & 7.28 & 11.60 & 60.44 & 35.06 \\
 & Ours &  & 82.51 & 26.44 & 40.62 & 49.81 & 44.29 & 60.50 & 11.37 & 14.61 & 58.87 & 38.31 \\
 & $\Delta$  & & \textbf{+1.35} & \textbf{+3.50} & \textbf{+6.87} & \textbf{+3.16} & \textbf{+1.53} & \textbf{+5.36} & \textbf{+4.09} & \textbf{+3.01} & -1.57 & \textbf{+3.26} \\
\bottomrule 
\end{tabular}
}
\vspace{-5pt}
\caption{Recall@1 on ImageNet and UnEd dataset for the baseline and our ViT-B/14 and ViT-L/14 models. Values are in percentage.}
\label{table:knn_eval}
\end{table*}

K-Nearest-Neighbor (KNN) task directly measures the performance of representations in image retrieval. This task generally involves a query and an index set, where queries are issued to retrieve the top relevant images from the index using the learned representation, and the metrics are defined by the similarity between the top retrieved candidates and the query image. We conduct a set of KNN evaluations on both ImageNet-1k and Universal Embedding Dataset (UnED) \cite{nikos2023uie}. For ImageNet-1k, we treat the validation split as the query set and the train split as the index set. For UnED, we follow the same split for each domain and the same preprocessing procedure for images described in \cite{nikos2023uie}. We use Recall@1 as the metric and report the results in Table \ref{table:knn_eval}.

On ImageNet KNN task, we observe consistent improvements over the baseline, which is expected as our models in general exhibit higher classification accuracy shown in Section \ref{subsec:classification}. On UnED dataset, the average performance for our models is significantly higher. In specific domains, such as cars and natural world images, our ImageNet-1k trained models show significantly improved performance (except for SOP, Inshop and Rp2k where the performance is degraded). Pre-training the model with ImageNet-21k gives performance improvements for nearly all datasets except Rp2k. Moreover, since the domains of these datasets do not necessarily overlap with ImageNet-1k, the improvement here also implies our model generalizes better in unseen classes.

\subsection{Semantic Segmentation}
\label{subsec:segmentation}

\begin{table}[h]
\centering
\scalebox{0.77}{
\begin{tabular}{c|c|c|cccc} 
\toprule
\multicolumn{2}{c|}{Model} & Training Setup &  ADE20k & P.Cont  & P.VOC & Citys. \\
\midrule
\multirow{9}*{\rotatebox[origin=c]{90}{ViT-B/14}} & Baseline& \multirow{3}*{IN-1k$_{300}$} & 43.27 & 49.86 & 73.27 & 65.83 \\
 & Ours  & & 43.89 & 50.71 & 73.40 & 67.46 \\
 & $\Delta$  & & \textbf{+0.61} & \textbf{+0.85} & \textbf{+0.13} & \textbf{+1.63} \\
\cmidrule{2-7}
 & Baseline& \multirow{3}*{IN-21k$_{30}$ FT} & 44.89 & 52.21 & 74.49 & 65.59 \\
 & Ours  & & 46.23 & 53.04 & 74.62 & 67.51 \\
 & $\Delta$  & & \textbf{+1.34} & \textbf{+0.83} & \textbf{+0.13} & \textbf{+1.92} \\
\cmidrule{2-7}
 & Baseline& \multirow{3}*{IN-21k$_{300}$ FT} & 45.59 & 52.85 & 75.09 & 66.37 \\
 & Ours  & & 46.80 & 53.45 & 75.65 & 67.82 \\
 & $\Delta$  & & \textbf{+1.21} & \textbf{+0.60} & \textbf{+0.56} & \textbf{+1.45} \\
\midrule
\multirow{6}*{\rotatebox[origin=c]{90}{ViT-L/14}}  & Baseline & \multirow{3}*{IN-21k$_{30}$ FT} & 46.30 & 53.60 & 75.19 & 68.64 \\
  & Ours & & 47.01 & 53.98 & 75.73 & 67.97 \\
 & $\Delta$  & & \textbf{+0.71} & \textbf{+0.38} & \textbf{+0.54} & -0.67 \\
\cmidrule{2-7}
  & Baseline & \multirow{3}*{IN-21k$_{300}$ FT} & 47.64 & 53.73 & 75.35 & 66.76 \\
 & Ours &  & 48.28 & 54.33 & 75.76 & 68.60 \\
 & $\Delta$  & & \textbf{+0.64} & \textbf{+0.60} & \textbf{+0.41} & \textbf{+1.84} \\
\bottomrule 
\end{tabular}
}
\vspace{-5pt}
\caption{Semantic segmentation evaluation (mIoU) on four datasets for the baseline and our ViT-B/14 and ViT-L/14 models. Numbers are in percentage.}
\label{table:segmentation_eval}
\end{table}

In this evaluation, we try to quantify the gains from the MIM self-supervised learning task introduced in our framework. The intuition is that a task such as MIM allows for better pixel-level understanding. We follow the same procedure described in \cite{caron2023locationaware}, which trains a linear decoder on semantic segmentation benchmarks on a higher resolution of $448\times 448$. Since patch size is the same, we perform a 2D bilinear upsampling of the positional embeddings to match the longer sequence length.
Here we pick four datasets: ADE20k \cite{zhou2017ade20k}, Pascal Context (``P.Cont”) \cite{mottaghi2014pcont}, Pascal VOC (``P.VOC”) \cite{everingham2010pvoc}, Cityscapes (``Citys.”) \cite{cordts2016citys}, and sweep over 7 different learning rates \{5e-4, 3e-4, 1e-4, 7e-5, 5e-3, 3e-5, 1e-5\} and 3 random seeds for each. We report best performing mean IoU metric on the validation datasets in Table \ref{table:segmentation_eval}.

We see in all four evaluations the model's performance is significantly improved. This demonstrates the MIM task enhances the learned information at patch level. In this regard, our setup can be viewed as a fusion strategy that encodes both global and local information in the representations.

\subsection{Ablation Study}
\label{subsec:ablation}
In this section, we show comprehensive analysis of ablations on two key design choices: choice of masking ratio and number of transformer layers in the decoder, and using \texttt{[CLS]} token instead of GAP in either classification or MIM task. We also ablate the impact of MIM task in pre-training and fine-tuning. In addition, we conduct a detailed comparison between our approach and a similar setup to DropToken \cite{hou2022droptoken}, where the same set of image patches are also dropped during supervised training. Unless explicitly mentioned, the ablation studies in this section are conducted on ViT-B/14 models.

\textbf{Impact of Masking Ratio and Decoder Size.} 
We evaluate the effect of varying masking ratios for the MIM task as well as the number of transformer layers in the decoder. We sweep the masking ratio between $5\%$ to $95\%$ with a $5\%$ interval. The choices for the number of decoder layers are \{1, 2, 4, 6, 8, 10, 12\}.
\begin{figure}[h]
\begin{center}
   \includegraphics[width=0.9\linewidth]{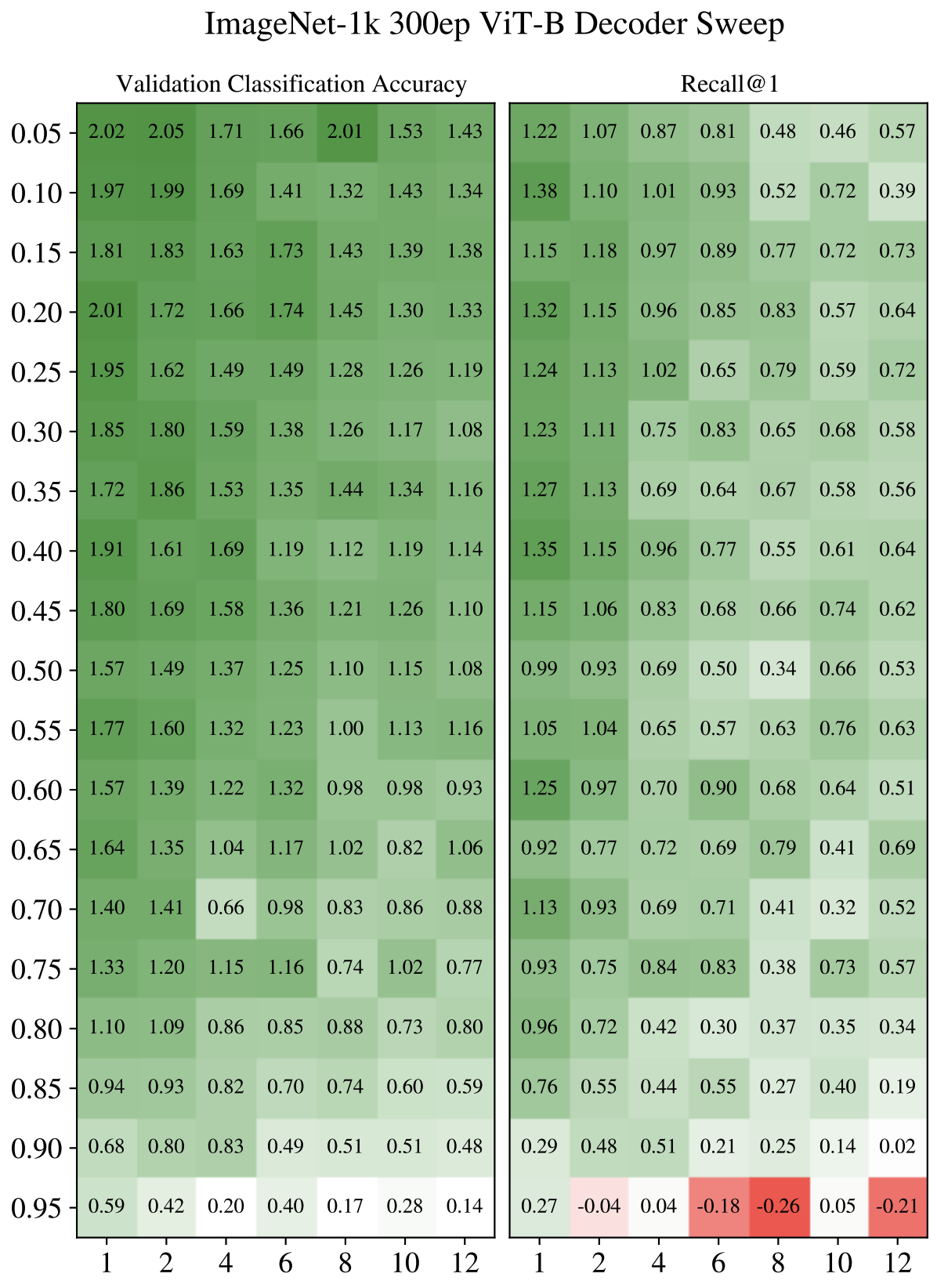}
\end{center}
\vspace{-5pt}
   \caption{The differences on ImageNet-1k validation accuracy (left table) and ImageNet KNN@1 (right table) relative to the baseline ViT-B/14 model for different masking ratio (y-axis) and number of decoder layers (x-axis), after training on ImageNet-1k for 300 epochs. Positive values, highlighted in different shades of green, indicate gains over the baseline performance. Red values denote the opposite. Baseline values are presented in Tables \ref{table:classification_eval} and \ref{table:knn_eval}.}
\label{fig:fig3}
\end{figure}

We show the difference in ImageNet-1k validation accuracy and KNN evaluation on the left and right panels in Figure \ref{fig:fig3}, respectively. The results demonstrate that for most of the hyparameter combinations, the validation accuracy and KNN performance are improved upon the baseline, indicating that our setup is very robust against masking ratio and the decoder size. Only at the very high 95\% masking ratio do we see the KNN metric suffer a little. Moreover, a small decoder consistently leads to better results regardless of masking ratio. This is likely due to forcing the encoder to learn more of the MIM task on its own rather than rely more on the decoder. Interestingly, unlike \cite{kaiming2022CVPR}, we see that lower masking ratios works better. We hypothesize that this is due to tasks being co-trained, so a lower masking ratio for MIM is more aligned with the classification task, which does not use any masking. %

\textbf{Impact of MIM Task in Pre-training and Fine-tuning.} 
In this study we study both ViT-B and ViT-L models with \texttt{IN-21k$_{300}$FT} setting and compare the classification performance by ablating MIM task in either pre-training or fine-tuning stage. We summarize ImageNet-1k, ImageNet-Real and ImageNet-v2 accuracy in Table \ref{table:mim_impact}. For both ViT-B and ViT-L models, we find that having MIM task in both pre-training and fine-tuning is essential for optimal performance. Having MIM only in pre-training can still result in better model performance than training without MIM task. However, using MIM task in fine-tuning alone leads to worse performance than the baseline. 

\iffalse
\begin{table}[h]
\centering
\scalebox{0.65}{
\begin{tabular}{c|c|c|c|c|c} 
\toprule
 & \makecell{MIM in \\Pre-training} & \makecell{MIM in \\Fine-tuning} & ImageNet-1k & ImageNet-Real & ImageNet-v2 \\
\midrule
\multirow{4}*{\rotatebox[origin=c]{90}{ViT-B/14}} & \ding{51} & \ding{51} & \textbf{83.68} (\uparrow 0.86) & \textbf{87.61} (\uparrow 1.00) & \textbf{72.15} (\uparrow 1.32)\\
  &  \ding{51} & \ding{56} & 83.59 (\uparrow 0.77) & 87.56 (\uparrow 0.95) & 71.73 (\uparrow 0.90) \\
  &  \ding{56} & \ding{51} & 82.55 (\downarrow 0.27) & 86.54 (\downarrow 0.07) & 70.20 (\downarrow 0.63) \\
  &  \ding{56} & \ding{56} & 82.82 & 86.61 & 70.83 \\
\midrule
\multirow{4}*{\rotatebox[origin=c]{90}{ViT-L/14}} & \ding{51} & \ding{51} & \textbf{83.57} (\uparrow 1.28)&  \textbf{87.11} (\uparrow 1.44) & \textbf{71.99} (\uparrow 2.15) \\
  &  \ding{51} & \ding{56} & 83.46 (\uparrow 1.17) & 87.10 (\uparrow 1.43) & 71.74 (\uparrow 1.90) \\
  &  \ding{56} & \ding{51} & 82.21 (\downarrow 0.08) & 85.67 (\uparrow 0.00) & 69.79 (\downarrow 0.05)\\
  &  \ding{56} & \ding{56} & 82.29 & 85.67 & 69.84 \\
\bottomrule 
\end{tabular}
}
\vspace{-5pt}
\caption{Classification benchmarks for ViT-B/14 and ViT-L/14 models trained with and without MIM task in ImageNet-21k pre-training and ImageNet-1k fine-tuning stages.}
\label{table:mim_impact}
\end{table}

\else

\begin{table}[h]
\centering
\scalebox{0.65}{
\begin{tabular}{c|c|c|c|c|c} 
\toprule
 & \makecell{MIM in \\Pre-training} & \makecell{MIM in \\Fine-tuning} & ImageNet-1k & ImageNet-Real & ImageNet-v2 \\
\midrule
\multirow{4}*{\rotatebox[origin=c]{90}{ViT-B/14}} & \ding{51} & \ding{51} & \textbf{83.68} (+0.86) & \textbf{87.61} (+1.00) & \textbf{72.15} (+1.32)\\
  &  \ding{51} & \ding{56} & 83.59 (+0.77) & 87.56 (+0.95) & 71.73 (+0.90) \\
  &  \ding{56} & \ding{51} & 82.55 (-0.27) & 86.54 (-0.07) & 70.20 (-0.63) \\
  &  \ding{56} & \ding{56} & 82.82 & 86.61 & 70.83 \\
\midrule
\multirow{4}*{\rotatebox[origin=c]{90}{ViT-L/14}} & \ding{51} & \ding{51} & \textbf{83.57} (+1.28)&  \textbf{87.11} (+1.44) & \textbf{71.99} (+2.15) \\
  &  \ding{51} & \ding{56} & 83.46 (+1.17) & 87.10 (+1.43) & 71.74 (+1.90) \\
  &  \ding{56} & \ding{51} & 82.21 (-0.08) & 85.67 (+0.00) & 69.79 (-0.05)\\
  &  \ding{56} & \ding{56} & 82.29 & 85.67 & 69.84 \\
\bottomrule 
\end{tabular}
}
\vspace{-5pt}
\caption{Classification benchmarks for ViT-B/14 and ViT-L/14 models trained with and without MIM task in ImageNet-21k pre-training and ImageNet-1k fine-tuning stages.}
\label{table:mim_impact}
\end{table}
\fi

\begin{figure}[h]
\begin{center}
   \includegraphics[width=1.0\linewidth]{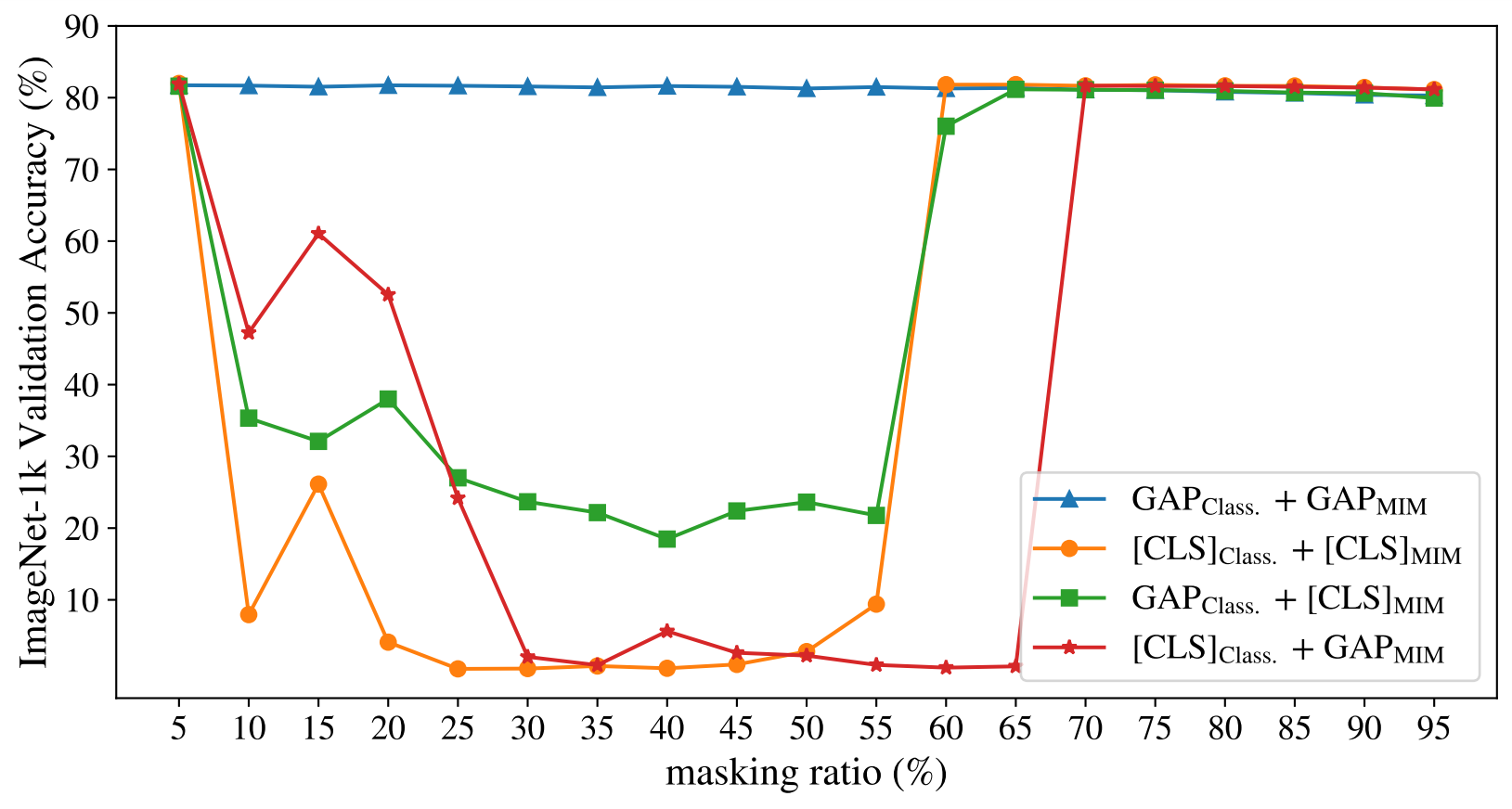}
\end{center}
\vspace{-5pt}
   \caption{ImageNet-1k validation accuracy (y-axis) for models trained with different masking ratios (x-axis). Four different configurations that use either GAP or \texttt{[CLS]} for classification task or the MIM task are shown in curves with different colors, and GAP$_{\text{Class.}}$ + GAP$_{\text{MIM}}$ corresponds to our choice of design. Better view in color.}
\label{fig:fig4}
\end{figure}

\textbf{Using \texttt{[CLS]} Token.}
In addition to our design choice of using GAP for both classification and MIM, denoted as GAP$_{\text{Class.}}$ + GAP$_{\text{MIM}}$, a learnable \texttt{[CLS]} token pre-pended to the patch sequence can be also considered in both tasks. Here we ablate using \texttt{[CLS]} in both tasks, which includes another $3$ design options: (1) \texttt{[CLS]} for both classification and MIM, denoted as \texttt{[CLS]}$_{\text{Class.}}$ + \texttt{[CLS]}$_{\text{MIM}}$, (2) \texttt{[CLS]} for classification and GAP for MIM, denoted as \texttt{[CLS]}$_{\text{Class.}}$ + GAP$_{\text{MIM}}$ and (3) GAP for classification and \texttt{[CLS]} for MIM, denoted as GAP$_{\text{Class.}}$ + \texttt{[CLS]}$_{\text{MIM}}$. Note when using \texttt{[CLS]} for the MIM task, at the input to the decoder in Figure \ref{fig:fig2}, the \texttt{[E]} token is instead replace by \texttt{[CLS]}. Using \texttt{[CLS]} for classification follows the standard setup that the classifier is applied on top of the encoded \texttt{[CLS]} token. For cases where both \texttt{[CLS]} and GAP are used (second and third configurations), we exclude \texttt{[CLS]} in the pooling. For these three options, the input sequence is longer than ours by $1$, resulting in a slightly higher baseline ImageNet-1k validation accuracy of $80.62\%$. We fix the decoder to have only one transformer layer and show the ImageNet-1k validation accuracy with respect to masking ratio for these three configurations together with our design in Figure ~\ref{fig:fig4}.

The configuration of \texttt{[CLS]}$_{\text{Class.}}$ + GAP$_{\text{MIM}}$ reaches the highest performance of $81.98\%$ accuracy at $5\%$ masking ratio. However all configurations that introduce \texttt{[CLS]} token in either task become unstable at intermediate masking ratios with significant drops in final accuracy metrics. Only at high masking ratios can the accuracy values recover back to the expected range. On the contrary, models training with GAP for both tasks show strong robustness across all masking setups.

\begin{figure}[h]
\begin{center}
   \includegraphics[width=1.0\linewidth]{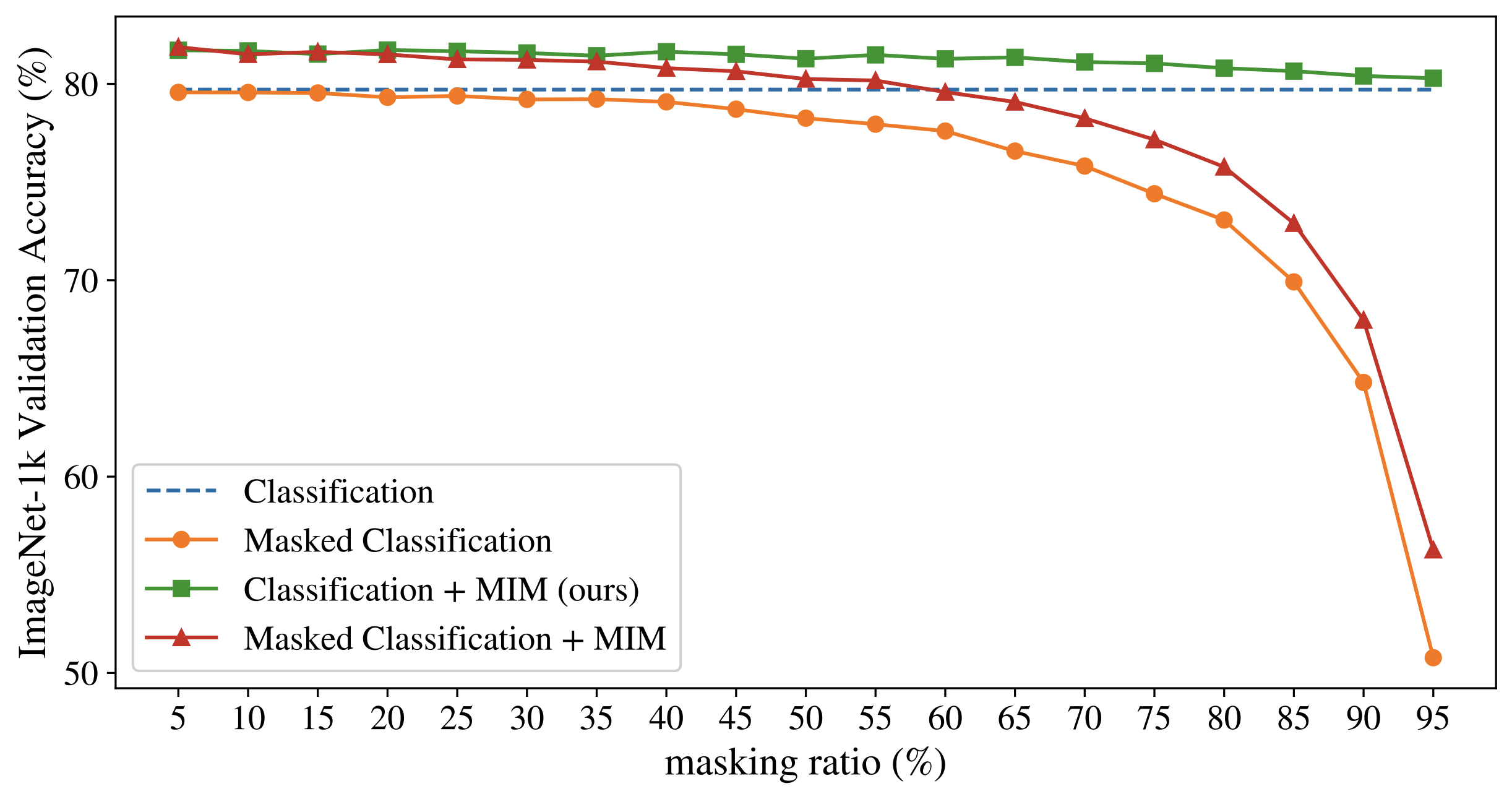}
\end{center}
\vspace{-5pt}
   \caption{The ImageNet-1k validation accuracy (y-axis) for models trained with different masking ratios (x-axis) when adding the same mask on the images in the classification task. Baseline models trained with masks in the classification task are shown in the orange curve, and the baseline model trained without any masking in classification is plotted in blue line. Better view in color.}
\label{fig:fig5}
\end{figure}

\textbf{Masking Images in Classification Training.} In our proposed setup we only apply a mask to the image in the MIM task, in principle for the classification task it is also possible to mask the input images. Similar ideas have been proved feasible in CLIP pre-training \cite{li2023cvpr} and Bert pre-training \cite{hou2022droptoken}. In this ablation study, we apply the same mask to both tasks. Concretely, supervised learning task now shares the same masked input as the MIM task. For completeness, we also include the baseline model trained with masked input in the comparison. We fix the number of transformer layers in the decoder to 1 and only study the effect of masking ratio on ImageNet-1k validation accuracy. The results are shown in Figure \ref{fig:fig5}.

After introducing masking to the classification task, we see a gradual decrease in accuracy when masking ratio increases, with or without the MIM task. This is expected since highly masked images might lose important semantic details that are crucial for classification. When training classification with masked images, adding MIM task still greatly improves the performance of the model, as shown by comparing the red and orange curves in Figure \ref{fig:fig5}. Another interesting observation is that, in the case of using masked images for both classification and MIM tasks, at lower masking ratio, the resulting model performance is still better than the vanilla baseline model trained with unmasked images (dashed blue curve). More specifically, at $50\%$ masking ratio, the former can still achieve $80.25\%$ validation accuracy. This indicates that using half of the patches for classifying the whole image and predicting the masked half, is as effective as using the whole image for classification. In another word, the presence of MIM task can greatly help reduce the information needed for the classification task to reach model quality parity.

\section{Conclusion}
\label{sec:conclusion}
In this work, we have presented a simple and very effective setup that combines supervised representation learning and MIM into one training stage. Our design involves a pre-trained visual tokenizer, and the addition of a shallow transformer-based decoder to the encoded representation. The MIM task predicts the highly semantic tokens in addition to the supervised classification task. We find that the introduction of this MIM task consistently improves model performance in classification, retrieval and semantic segmentation while preserving strong robustness against different masking ratios. We also show the effectiveness of our method on large-scale pre-training and fine-tuning and larger model sizes. We conduct detailed ablations on the choice of hyperparameters and comparisons between different design options.

{
    \small
    \bibliographystyle{ieeenat_fullname}
    \bibliography{main}

\begin{thebibliography}{64}
\providecommand{\natexlab}[1]{#1}
\providecommand{\url}[1]{\texttt{#1}}
\expandafter\ifx\csname urlstyle\endcsname\relax
  \providecommand{\doi}[1]{doi: #1}\else
  \providecommand{\doi}{doi: \begingroup \urlstyle{rm}\Url}\fi

\bibitem[Abadi et~al.(2015)Abadi, Agarwal, Barham, Brevdo, Chen, Citro,
  Corrado, Davis, Dean, Devin, Ghemawat, Goodfellow, Harp, Irving, Isard, Jia,
  Jozefowicz, Kaiser, Kudlur, Levenberg, Man\'{e}, Monga, Moore, Murray, Olah,
  Schuster, Shlens, Steiner, Sutskever, Talwar, Tucker, Vanhoucke, Vasudevan,
  Vi\'{e}gas, Vinyals, Warden, Wattenberg, Wicke, Yu, and
  Zheng]{tensorflow2015-whitepaper}
Mart\'{i}n Abadi, Ashish Agarwal, Paul Barham, Eugene Brevdo, Zhifeng Chen,
  Craig Citro, Greg~S. Corrado, Andy Davis, Jeffrey Dean, Matthieu Devin,
  Sanjay Ghemawat, Ian Goodfellow, Andrew Harp, Geoffrey Irving, Michael Isard,
  Yangqing Jia, Rafal Jozefowicz, Lukasz Kaiser, Manjunath Kudlur, Josh
  Levenberg, Dandelion Man\'{e}, Rajat Monga, Sherry Moore, Derek Murray, Chris
  Olah, Mike Schuster, Jonathon Shlens, Benoit Steiner, Ilya Sutskever, Kunal
  Talwar, Paul Tucker, Vincent Vanhoucke, Vijay Vasudevan, Fernanda Vi\'{e}gas,
  Oriol Vinyals, Pete Warden, Martin Wattenberg, Martin Wicke, Yuan Yu, and
  Xiaoqiang Zheng.
\newblock {TensorFlow}: Large-scale machine learning on heterogeneous systems,
  2015.
\newblock Software available from tensorflow.org.

\bibitem[Bao et~al.(2022)Bao, Dong, Piao, and Wei]{bao2022beit}
Hangbo Bao, Li Dong, Songhao Piao, and Furu Wei.
\newblock {BE}it: {BERT} pre-training of image transformers.
\newblock In \emph{ICLR}, 2022.

\bibitem[Beyer et~al.(2020)Beyer, H{\'{e}}naff, Kolesnikov, Zhai, and van~den
  Oord]{imagenet_real}
Lucas Beyer, Olivier~J. H{\'{e}}naff, Alexander Kolesnikov, Xiaohua Zhai, and
  A{\"{a}}ron van~den Oord.
\newblock Are we done with imagenet?
\newblock \emph{CoRR}, abs/2006.07159, 2020.

\bibitem[Caron et~al.(2020)Caron, Misra, Mairal, Goyal, Bojanowski, and
  Joulin]{mathilde2020swav}
Mathilde Caron, Ishan Misra, Julien Mairal, Priya Goyal, Piotr Bojanowski, and
  Armand Joulin.
\newblock Unsupervised learning of visual features by contrasting cluster
  assignments.
\newblock In \emph{NeurIPS}, pages 9912--9924. Curran Associates, Inc., 2020.

\bibitem[Caron et~al.(2021)Caron, Touvron, Misra, J\'egou, Mairal, Bojanowski,
  and Joulin]{Caron2021dino}
Mathilde Caron, Hugo Touvron, Ishan Misra, Herv\'e J\'egou, Julien Mairal,
  Piotr Bojanowski, and Armand Joulin.
\newblock Emerging properties in self-supervised vision transformers.
\newblock In \emph{Proceedings of the IEEE/CVF International Conference on
  Computer Vision (ICCV)}, pages 9650--9660, 2021.

\bibitem[Caron et~al.(2023)Caron, Houlsby, and Schmid]{caron2023locationaware}
Mathilde Caron, Neil Houlsby, and Cordelia Schmid.
\newblock Location-aware self-supervised transformers for semantic
  segmentation, 2023.

\bibitem[Chang et~al.(2022)Chang, Zhang, Jiang, Liu, and
  Freeman]{Chang2022maskgit}
Huiwen Chang, Han Zhang, Lu Jiang, Ce Liu, and William~T. Freeman.
\newblock Maskgit: Masked generative image transformer.
\newblock In \emph{Proceedings of the IEEE/CVF Conference on Computer Vision
  and Pattern Recognition (CVPR)}, pages 11315--11325, 2022.

\bibitem[Chang et~al.(2023)Chang, Zhang, Barber, Maschinot, Lezama, Jiang,
  Yang, Murphy, Freeman, Rubinstein, Li, and Krishnan]{chang2023muse}
Huiwen Chang, Han Zhang, Jarred Barber, AJ Maschinot, Jose Lezama, Lu Jiang,
  Ming-Hsuan Yang, Kevin Murphy, William~T. Freeman, Michael Rubinstein,
  Yuanzhen Li, and Dilip Krishnan.
\newblock Muse: Text-to-image generation via masked generative transformers,
  2023.

\bibitem[Chen et~al.(2020)Chen, Kornblith, Norouzi, and Hinton]{2020simclr}
Ting Chen, Simon Kornblith, Mohammad Norouzi, and Geoffrey Hinton.
\newblock A simple framework for contrastive learning of visual
  representations.
\newblock pages 1597--1607. PMLR, 2020.

\bibitem[Chen et~al.(2023)Chen, Wang, Changpinyo, Piergiovanni, Padlewski,
  Salz, Goodman, Grycner, Mustafa, Beyer, Kolesnikov, Puigcerver, Ding, Rong,
  Akbari, Mishra, Xue, Thapliyal, Bradbury, Kuo, Seyedhosseini, Jia, Ayan,
  Ruiz, Steiner, Angelova, Zhai, Houlsby, and Soricut]{chen2023pali}
Xi Chen, Xiao Wang, Soravit Changpinyo, AJ Piergiovanni, Piotr Padlewski,
  Daniel Salz, Sebastian Goodman, Adam Grycner, Basil Mustafa, Lucas Beyer,
  Alexander Kolesnikov, Joan Puigcerver, Nan Ding, Keran Rong, Hassan Akbari,
  Gaurav Mishra, Linting Xue, Ashish~V Thapliyal, James Bradbury, Weicheng Kuo,
  Mojtaba Seyedhosseini, Chao Jia, Burcu~Karagol Ayan, Carlos~Riquelme Ruiz,
  Andreas~Peter Steiner, Anelia Angelova, Xiaohua Zhai, Neil Houlsby, and Radu
  Soricut.
\newblock Pa{LI}: A jointly-scaled multilingual language-image model.
\newblock In \emph{ICLR}, 2023.

\bibitem[Cheng et~al.(2017)Cheng, Han, and Lu]{Resisc45}
Gong Cheng, Junwei Han, and Xiaoqiang Lu.
\newblock Remote sensing image scene classification: Benchmark and state of the
  art.
\newblock \emph{Proceedings of the IEEE}, 105\penalty0 (10):\penalty0
  1865--1883, 2017.

\bibitem[Chopra et~al.(2005)Chopra, Hadsell, and LeCun]{lecun2005}
S. Chopra, R. Hadsell, and Y. LeCun.
\newblock Learning a similarity metric discriminatively, with application to
  face verification.
\newblock In \emph{CVPR}, pages 539--546 vol. 1, 2005.

\bibitem[Cordts et~al.(2016)Cordts, Omran, Ramos, Rehfeld, Enzweiler, Benenson,
  Franke, Roth, and Schiele]{cordts2016citys}
Marius Cordts, Mohamed Omran, Sebastian Ramos, Timo Rehfeld, Markus Enzweiler,
  Rodrigo Benenson, Uwe Franke, Stefan Roth, and Bernt Schiele.
\newblock The cityscapes dataset for semantic urban scene understanding.
\newblock In \emph{CVPR}, 2016.

\bibitem[Dehghani et~al.(2022)Dehghani, Gritsenko, Arnab, Minderer, and
  Tay]{dehghani2021scenic}
Mostafa Dehghani, Alexey Gritsenko, Anurag Arnab, Matthias Minderer, and Yi
  Tay.
\newblock Scenic: A jax library for computer vision research and beyond.
\newblock In \emph{CVPR}, pages 21393--21398, 2022.

\bibitem[Deng et~al.(2009)Deng, Dong, Socher, Li, Li, and
  Fei-Fei]{degn2009imagenet}
Jia Deng, Wei Dong, Richard Socher, Li-Jia Li, Kai Li, and Li Fei-Fei.
\newblock Imagenet: A large-scale hierarchical image database.
\newblock In \emph{CVPR}, pages 248--255, 2009.

\bibitem[Deng et~al.(2019)Deng, Guo, Xue, and Zafeiriou]{2019arcface}
Jiankang Deng, Jia Guo, Niannan Xue, and Stefanos Zafeiriou.
\newblock Arcface: Additive angular margin loss for deep face recognition.
\newblock In \emph{CVPR}, 2019.

\bibitem[Devlin et~al.(2019)Devlin, Chang, Lee, and Toutanova]{2019bert}
Jacob Devlin, Ming-Wei Chang, Kenton Lee, and Kristina Toutanova.
\newblock {BERT}: Pre-training of deep bidirectional transformers for language
  understanding.
\newblock In \emph{North {A}merican Chapter of the Association for
  Computational Linguistics}, pages 4171--4186, Minneapolis, Minnesota, 2019.
  Association for Computational Linguistics.

\bibitem[Dong et~al.(2023)Dong, Bao, Zhang, Chen, Zhang, Yuan, Chen, Wen, Yu,
  and Guo]{peco2023}
Xiaoyi Dong, Jianmin Bao, Ting Zhang, Dongdong Chen, Weiming Zhang, Lu Yuan,
  Dong Chen, Fang Wen, Nenghai Yu, and Baining Guo.
\newblock Peco: Perceptual codebook for bert pre-training of vision
  transformers.
\newblock \emph{AAAI}, 37\penalty0 (1):\penalty0 552--560, 2023.

\bibitem[Esser et~al.(2021)Esser, Rombach, and Ommer]{esser2021vqgan}
Patrick Esser, Robin Rombach, and Bjorn Ommer.
\newblock Taming transformers for high-resolution image synthesis.
\newblock In \emph{CVPR}, pages 12873--12883, 2021.

\bibitem[Everingham et~al.(2010)Everingham, Van~Gool, Williams, Winn, and
  Zisserman]{everingham2010pvoc}
Mark Everingham, Luc Van~Gool, Christopher K.~I. Williams, John Winn, and
  Andrew Zisserman.
\newblock The pascal visual object classes (voc) challenge.
\newblock \emph{IJCV}, 88\penalty0 (2):\penalty0 303--338, 2010.

\bibitem[Geiger et~al.(2013)Geiger, Lenz, Stiller, and Urtasun]{Kitti-distance}
Andreas Geiger, P Lenz, Christoph Stiller, and Raquel Urtasun.
\newblock Vision meets robotics: the kitti dataset.
\newblock \emph{The International Journal of Robotics Research}, 32:\penalty0
  1231--1237, 2013.

\bibitem[Grill et~al.(2020)Grill, Strub, Altch\'{e}, Tallec, Richemond,
  Buchatskaya, Doersch, Avila~Pires, Guo, Gheshlaghi~Azar, Piot, kavukcuoglu,
  Munos, and Valko]{grill2020boyl}
Jean-Bastien Grill, Florian Strub, Florent Altch\'{e}, Corentin Tallec, Pierre
  Richemond, Elena Buchatskaya, Carl Doersch, Bernardo Avila~Pires, Zhaohan
  Guo, Mohammad Gheshlaghi~Azar, Bilal Piot, koray kavukcuoglu, Remi Munos, and
  Michal Valko.
\newblock Bootstrap your own latent - a new approach to self-supervised
  learning.
\newblock In \emph{NeurIPS}, pages 21271--21284. Curran Associates, Inc., 2020.

\bibitem[He et~al.(2016)He, Zhang, Ren, and Sun]{he2016resnet}
Kaiming He, Xiangyu Zhang, Shaoqing Ren, and Jian Sun.
\newblock Deep residual learning for image recognition.
\newblock In \emph{Proceedings of the IEEE Conference on Computer Vision and
  Pattern Recognition (CVPR)}, 2016.

\bibitem[He et~al.(2020)He, Fan, Wu, Xie, and Girshick]{he2020moco}
Kaiming He, Haoqi Fan, Yuxin Wu, Saining Xie, and Ross Girshick.
\newblock Momentum contrast for unsupervised visual representation learning.
\newblock In \emph{CVPR}, 2020.

\bibitem[He et~al.(2022)He, Chen, Xie, Li, Doll\'ar, and
  Girshick]{kaiming2022CVPR}
Kaiming He, Xinlei Chen, Saining Xie, Yanghao Li, Piotr Doll\'ar, and Ross
  Girshick.
\newblock Masked autoencoders are scalable vision learners.
\newblock In \emph{CVPR}, pages 16000--16009, 2022.

\bibitem[Hoffer and Ailon(2015)]{tripletloss}
Elad Hoffer and Nir Ailon.
\newblock Deep metric learning using triplet network.
\newblock In \emph{Similarity-Based Pattern Recognition}, pages 84--92, Cham,
  2015. Springer International Publishing.

\bibitem[Hou et~al.(2022)Hou, Pang, Zhou, Wu, Song, Song, and
  Zhou]{hou2022droptoken}
Le Hou, Richard~Yuanzhe Pang, Tianyi Zhou, Yuexin Wu, Xinying Song, Xiaodan
  Song, and Denny Zhou.
\newblock Token dropping for efficient {BERT} pretraining.
\newblock In \emph{Proceedings of the 60th Annual Meeting of the Association
  for Computational Linguistics (Volume 1: Long Papers)}, pages 3774--3784,
  Dublin, Ireland, 2022. Association for Computational Linguistics.

\bibitem[Hu et~al.(2018)Hu, Wang, Yang, Komlev, Huang, Chen, Huang, Wu,
  Merchant, and Sacheti]{2018bingsearch}
Houdong Hu, Yan Wang, Linjun Yang, Pavel Komlev, Li Huang, Xi~(Stephen) Chen,
  Jiapei Huang, Ye Wu, Meenaz Merchant, and Arun Sacheti.
\newblock Web-scale responsive visual search at bing.
\newblock In \emph{Proceedings of the 24th ACM SIGKDD International Conference
  on Knowledge Discovery \& Data Mining}, page 359–367, New York, NY, USA,
  2018. Association for Computing Machinery.

\bibitem[Jing et~al.(2015)Jing, Liu, Kislyuk, Zhai, Xu, Donahue, and
  Tavel]{2015pinterest}
Yushi Jing, David Liu, Dmitry Kislyuk, Andrew Zhai, Jiajing Xu, Jeff Donahue,
  and Sarah Tavel.
\newblock Visual search at pinterest.
\newblock In \emph{Proceedings of the 21th ACM SIGKDD International Conference
  on Knowledge Discovery and Data Mining}, page 1889–1898, New York, NY, USA,
  2015. Association for Computing Machinery.

\bibitem[Kirillov et~al.(2023)Kirillov, Mintun, Ravi, Mao, Rolland, Gustafson,
  Xiao, Whitehead, Berg, Lo, Dollar, and Girshick]{kirillov2023sam}
Alexander Kirillov, Eric Mintun, Nikhila Ravi, Hanzi Mao, Chloe Rolland, Laura
  Gustafson, Tete Xiao, Spencer Whitehead, Alexander~C. Berg, Wan-Yen Lo, Piotr
  Dollar, and Ross Girshick.
\newblock Segment anything.
\newblock In \emph{ICCV}, pages 4015--4026, 2023.

\bibitem[Kolesnikov et~al.(2021)Kolesnikov, Dosovitskiy, Weissenborn, Heigold,
  Uszkoreit, Beyer, Minderer, Dehghani, Houlsby, Gelly, Unterthiner, and
  Zhai]{alexander2021vit}
Alexander Kolesnikov, Alexey Dosovitskiy, Dirk Weissenborn, Georg Heigold,
  Jakob Uszkoreit, Lucas Beyer, Matthias Minderer, Mostafa Dehghani, Neil
  Houlsby, Sylvain Gelly, Thomas Unterthiner, and Xiaohua Zhai.
\newblock An image is worth 16x16 words: Transformers for image recognition at
  scale.
\newblock In \emph{ICLR}, 2021.

\bibitem[Krizhevsky et~al.(2012)Krizhevsky, Sutskever, and
  Hinton]{NIPS2012_c399862d}
Alex Krizhevsky, Ilya Sutskever, and Geoffrey~E Hinton.
\newblock Imagenet classification with deep convolutional neural networks.
\newblock In \emph{Advances in Neural Information Processing Systems}. Curran
  Associates, Inc., 2012.

\bibitem[Li et~al.(2023{\natexlab{a}})Li, Li, and Hoi]{li2023blipdiffusion}
Dongxu Li, Junnan Li, and Steven C.~H. Hoi.
\newblock Blip-diffusion: Pre-trained subject representation for controllable
  text-to-image generation and editing, 2023{\natexlab{a}}.

\bibitem[Li et~al.(2023{\natexlab{b}})Li, Chang, Mishra, Zhang, Katabi, and
  Krishnan]{Li2023mage}
Tianhong Li, Huiwen Chang, Shlok Mishra, Han Zhang, Dina Katabi, and Dilip
  Krishnan.
\newblock Mage: Masked generative encoder to unify representation learning and
  image synthesis.
\newblock In \emph{CVPR}, pages 2142--2152, 2023{\natexlab{b}}.

\bibitem[Li et~al.(2023{\natexlab{c}})Li, Fan, Hu, Feichtenhofer, and
  He]{li2023cvpr}
Yanghao Li, Haoqi Fan, Ronghang Hu, Christoph Feichtenhofer, and Kaiming He.
\newblock Scaling language-image pre-training via masking.
\newblock In \emph{CVPR}, pages 23390--23400, 2023{\natexlab{c}}.

\bibitem[Liu et~al.(2017)Liu, Wen, Yu, Li, Raj, and Song]{liu2017sphereface}
Weiyang Liu, Yandong Wen, Zhiding Yu, Ming Li, Bhiksha Raj, and Le Song.
\newblock Sphereface: Deep hypersphere embedding for face recognition.
\newblock In \emph{CVPR}, 2017.

\bibitem[Loshchilov and Hutter(2019)]{loshchilov2018adamw}
Ilya Loshchilov and Frank Hutter.
\newblock Decoupled weight decay regularization.
\newblock In \emph{ICLR}, 2019.

\bibitem[Mottaghi et~al.(2014)Mottaghi, Chen, Liu, Cho, Lee, Fidler, Urtasun,
  and Yuille]{mottaghi2014pcont}
Roozbeh Mottaghi, Xianjie Chen, Xiaobai Liu, Nam-Gyu Cho, Seong-Whan Lee, Sanja
  Fidler, Raquel Urtasun, and Alan Yuille.
\newblock The role of context for object detection and semantic segmentation in
  the wild.
\newblock In \emph{CVPR}, 2014.

\bibitem[Oquab et~al.(2023)Oquab, Darcet, Moutakanni, Vo, Szafraniec, Khalidov,
  Fernandez, Haziza, Massa, El-Nouby, Assran, Ballas, Galuba, Howes, Huang, Li,
  Misra, Rabbat, Sharma, Synnaeve, Xu, Jegou, Mairal, Labatut, Joulin, and
  Bojanowski]{oquab2023dinov2}
Maxime Oquab, Timothée Darcet, Théo Moutakanni, Huy Vo, Marc Szafraniec,
  Vasil Khalidov, Pierre Fernandez, Daniel Haziza, Francisco Massa, Alaaeldin
  El-Nouby, Mahmoud Assran, Nicolas Ballas, Wojciech Galuba, Russell Howes,
  Po-Yao Huang, Shang-Wen Li, Ishan Misra, Michael Rabbat, Vasu Sharma, Gabriel
  Synnaeve, Hu Xu, Hervé Jegou, Julien Mairal, Patrick Labatut, Armand Joulin,
  and Piotr Bojanowski.
\newblock Dinov2: Learning robust visual features without supervision, 2023.

\bibitem[Parkhi et~al.(2012)Parkhi, Vedaldi, Zisserman, and
  Jawahar]{OxfordPets}
Omkar~M Parkhi, Andrea Vedaldi, Andrew Zisserman, and C.~V. Jawahar.
\newblock Cats and dogs.
\newblock In \emph{2012 IEEE Conference on Computer Vision and Pattern
  Recognition}, pages 3498--3505, 2012.

\bibitem[Peng et~al.(2022)Peng, Dong, Bao, Ye, and Wei]{peng2022beit}
Zhiliang Peng, Li Dong, Hangbo Bao, Qixiang Ye, and Furu Wei.
\newblock Beit v2: Masked image modeling with vector-quantized visual
  tokenizers, 2022.

\bibitem[Ramesh et~al.(2022)Ramesh, Dhariwal, Nichol, Chu, and
  Chen]{ramesh2022dalle2}
Aditya Ramesh, Prafulla Dhariwal, Alex Nichol, Casey Chu, and Mark Chen.
\newblock Hierarchical text-conditional image generation with clip latents,
  2022.

\bibitem[Recht et~al.(2019)Recht, Roelofs, Schmidt, and Shankar]{imagenet_v2}
Benjamin Recht, Rebecca Roelofs, Ludwig Schmidt, and Vaishaal Shankar.
\newblock Do {I}mage{N}et classifiers generalize to {I}mage{N}et?
\newblock In \emph{Proceedings of the 36th International Conference on Machine
  Learning}, pages 5389--5400. PMLR, 2019.

\bibitem[Schroff et~al.(2015)Schroff, Kalenichenko, and
  Philbin]{schroff2014facenet}
Florian Schroff, Dmitry Kalenichenko, and James Philbin.
\newblock Facenet: A unified embedding for face recognition and clustering.
\newblock In \emph{CVPR}, 2015.

\bibitem[Shao et~al.(2023)Shao, Chen, Karpur, Cui, Araujo, and Cao]{shao2023}
Shihao Shao, Kaifeng Chen, Arjun Karpur, Qinghua Cui, Andr\'e Araujo, and
  Bingyi Cao.
\newblock Global features are all you need for image retrieval and reranking.
\newblock In \emph{ICCV}, pages 11036--11046, 2023.

\bibitem[Shi et~al.(2023)Shi, Xiong, Lin, and Jung]{shi2023instantbooth}
Jing Shi, Wei Xiong, Zhe Lin, and Hyun~Joon Jung.
\newblock Instantbooth: Personalized text-to-image generation without test-time
  finetuning, 2023.

\bibitem[Sohn(2016)]{kihyuk2016}
Kihyuk Sohn.
\newblock Improved deep metric learning with multi-class n-pair loss objective.
\newblock In \emph{NeurIPS}. Curran Associates, Inc., 2016.

\bibitem[Steiner et~al.(2022)Steiner, Kolesnikov, Zhai, Wightman, Uszkoreit,
  and Beyer]{steiner2022augreg}
Andreas~Peter Steiner, Alexander Kolesnikov, Xiaohua Zhai, Ross Wightman, Jakob
  Uszkoreit, and Lucas Beyer.
\newblock How to train your vit? data, augmentation, and regularization in
  vision transformers.
\newblock \emph{Transactions on Machine Learning Research}, 2022.

\bibitem[Touvron et~al.(2021)Touvron, Cord, Douze, Massa, Sablayrolles, and
  Jegou]{deit}
Hugo Touvron, Matthieu Cord, Matthijs Douze, Francisco Massa, Alexandre
  Sablayrolles, and Herve Jegou.
\newblock Training data-efficient image transformers; distillation through
  attention.
\newblock pages 10347--10357. PMLR, 2021.

\bibitem[Touvron et~al.(2022)Touvron, Cord, and J{\'e}gou]{hugo2022}
Hugo Touvron, Matthieu Cord, and Herv{\'e} J{\'e}gou.
\newblock Deit iii: Revenge of the vit.
\newblock In \emph{ECCV}, pages 516--533, Cham, 2022. Springer Nature
  Switzerland.

\bibitem[Vaswani et~al.(2017)Vaswani, Shazeer, Parmar, Uszkoreit, Jones, Gomez,
  Kaiser, and Polosukhin]{Vaswania2017attention}
Ashish Vaswani, Noam Shazeer, Niki Parmar, Jakob Uszkoreit, Llion Jones,
  Aidan~N Gomez, \L~ukasz Kaiser, and Illia Polosukhin.
\newblock Attention is all you need.
\newblock In \emph{NeurIPS}. Curran Associates, Inc., 2017.

\bibitem[Wang et~al.(2018{\natexlab{a}})Wang, Cheng, Liu, and
  Liu]{2018marginsoftmax}
Feng Wang, Jian Cheng, Weiyang Liu, and Haijun Liu.
\newblock Additive margin softmax for face verification.
\newblock \emph{IEEE Signal Processing Letters}, 25\penalty0 (7):\penalty0
  926--930, 2018{\natexlab{a}}.

\bibitem[Wang et~al.(2018{\natexlab{b}})Wang, Wang, Zhou, Ji, Gong, Zhou, Li,
  and Liu]{wang2018cosface}
Hao Wang, Yitong Wang, Zheng Zhou, Xing Ji, Dihong Gong, Jingchao Zhou, Zhifeng
  Li, and Wei Liu.
\newblock Cosface: Large margin cosine loss for deep face recognition.
\newblock In \emph{CVPR}, 2018{\natexlab{b}}.

\bibitem[Wang et~al.(2014)Wang, Song, Leung, Rosenberg, Wang, Philbin, Chen,
  and Wu]{wang2014finegrain}
Jiang Wang, Yang Song, Thomas Leung, Chuck Rosenberg, Jingbin Wang, James
  Philbin, Bo Chen, and Ying Wu.
\newblock Learning fine-grained image similarity with deep ranking.
\newblock In \emph{Proceedings of the IEEE Conference on Computer Vision and
  Pattern Recognition (CVPR)}, 2014.

\bibitem[Wang et~al.(2023)Wang, Bao, Dong, Bjorck, Peng, Liu, Aggarwal,
  Mohammed, Singhal, Som, and Wei]{wang2023beit3}
Wenhui Wang, Hangbo Bao, Li Dong, Johan Bjorck, Zhiliang Peng, Qiang Liu, Kriti
  Aggarwal, Owais~Khan Mohammed, Saksham Singhal, Subhojit Som, and Furu Wei.
\newblock Image as a foreign language: Beit pretraining for vision and
  vision-language tasks.
\newblock In \emph{CVPR}, pages 19175--19186, 2023.

\bibitem[Wei et~al.(2022{\natexlab{a}})Wei, Fan, Xie, Wu, Yuille, and
  Feichtenhofer]{wei2022maskfeat}
Chen Wei, Haoqi Fan, Saining Xie, Chao-Yuan Wu, Alan Yuille, and Christoph
  Feichtenhofer.
\newblock Masked feature prediction for self-supervised visual pre-training.
\newblock In \emph{CVPR}, pages 14668--14678, 2022{\natexlab{a}}.

\bibitem[Wei et~al.(2022{\natexlab{b}})Wei, Xie, Zhou, Li, and
  Tian]{wei2022mvp}
Longhui Wei, Lingxi Xie, Wengang Zhou, Houqiang Li, and Qi Tian.
\newblock Mvp: Multimodality-guided visual pre-training.
\newblock In \emph{ECCV}, pages 337--353, Cham, 2022{\natexlab{b}}. Springer
  Nature Switzerland.

\bibitem[Wightman et~al.(2021)Wightman, Touvron, and Jegou]{wightman2021resnet}
Ross Wightman, Hugo Touvron, and Herve Jegou.
\newblock Resnet strikes back: An improved training procedure in timm.
\newblock In \emph{NeurIPS 2021 Workshop on ImageNet: Past, Present, and
  Future}, 2021.

\bibitem[Ypsilantis et~al.(2023)Ypsilantis, Chen, Cao, Lipovsk\'y,
  Dogan-Sch\"onberger, Makosa, Bluntschli, Seyedhosseini, Chum, and
  Araujo]{nikos2023uie}
Nikolaos-Antonios Ypsilantis, Kaifeng Chen, Bingyi Cao, M\'ario Lipovsk\'y,
  Pelin Dogan-Sch\"onberger, Grzegorz Makosa, Boris Bluntschli, Mojtaba
  Seyedhosseini, Ond\v{r}ej Chum, and Andr\'e Araujo.
\newblock Towards universal image embeddings: A large-scale dataset and
  challenge for generic image representations.
\newblock In \emph{ICCV}, pages 11290--11301, 2023.

\bibitem[Yu et~al.(2022)Yu, Li, Koh, Zhang, Pang, Qin, Ku, Xu, Baldridge, and
  Wu]{yu2022vectorquantized}
Jiahui Yu, Xin Li, Jing~Yu Koh, Han Zhang, Ruoming Pang, James Qin, Alexander
  Ku, Yuanzhong Xu, Jason Baldridge, and Yonghui Wu.
\newblock Vector-quantized image modeling with improved {VQGAN}.
\newblock In \emph{ICLR}, 2022.

\bibitem[Zhai and Wu(2019)]{zhai2019classification}
Andrew Zhai and Hao-Yu Wu.
\newblock Classification is a strong baseline for deep metric learning, 2019.

\bibitem[Zhai et~al.(2020)Zhai, Puigcerver, Kolesnikov, Ruyssen, Riquelme,
  Lucic, Djolonga, Pinto, Neumann, Dosovitskiy, Beyer, Bachem, Tschannen,
  Michalski, Bousquet, Gelly, and Houlsby]{zhai2020vtab}
Xiaohua Zhai, Joan Puigcerver, Alexander Kolesnikov, Pierre Ruyssen, Carlos
  Riquelme, Mario Lucic, Josip Djolonga, Andre~Susano Pinto, Maxim Neumann,
  Alexey Dosovitskiy, Lucas Beyer, Olivier Bachem, Michael Tschannen, Marcin
  Michalski, Olivier Bousquet, Sylvain Gelly, and Neil Houlsby.
\newblock A large-scale study of representation learning with the visual task
  adaptation benchmark, 2020.

\bibitem[Zhai et~al.(2022)Zhai, Kolesnikov, Houlsby, and
  Beyer]{zhao2022scalingvit}
Xiaohua Zhai, Alexander Kolesnikov, Neil Houlsby, and Lucas Beyer.
\newblock Scaling vision transformers.
\newblock In \emph{CVPR}, pages 12104--12113, 2022.

\bibitem[Zhou et~al.(2017)Zhou, Zhao, Puig, Fidler, Barriuso, and
  Torralba]{zhou2017ade20k}
Bolei Zhou, Hang Zhao, Xavier Puig, Sanja Fidler, Adela Barriuso, and Antonio
  Torralba.
\newblock Scene parsing through ade20k dataset.
\newblock In \emph{CVPR}, 2017.

\end{thebibliography}
}

\newpage

\makeatletter
\renewcommand{\thefigure}{S\@arabic\c@figure}
\renewcommand{\thetable}{S\@arabic\c@table}
\makeatletter

\setcounter{figure}{0}
\setcounter{table}{0}

\appendix
\clearpage
\setcounter{page}{1}
\maketitlesupplementary
\section{Full VTAB Results}
\label{sec:full_vtab}

We evaluate the transfer capability of our method on 4 main computer vision datasets from the VTAB benchmark \cite{zhai2020vtab}: CIFAR-100 \cite{NIPS2012_c399862d}, Oxford IIIT Pets \cite{OxfordPets}, Resisc45 \cite{Resisc45} and Kitti-distance \cite{Kitti-distance}. We also report the mean value across all 19 VTAB benchmark datasets. The full VTAB evaluation results are in Table~\ref{table:vtab_eval_full}. 

For all VTAB datasets we finetune at a higher resolution of $384$ and sweep over $4$ different learning rates \{1e-3, 3e-3, 1e-2, 3e-2\} and 2 combinations of total training and warmup steps \{(500, 100), (2500, 200)\}. The validation sets are chosen as in \cite{steiner2022augreg}. We report final numbers on top-1 classification accuracy based on highest performing hyperparameters using validation accuracy, as shown in Table \ref{table:vtab_eval_full}.

\section{MIM Loss on Masked Tokens}
\label{sec:masked_token_loss}
\begin{figure*}[t]
\begin{center}
   \includegraphics[width=0.9\linewidth]{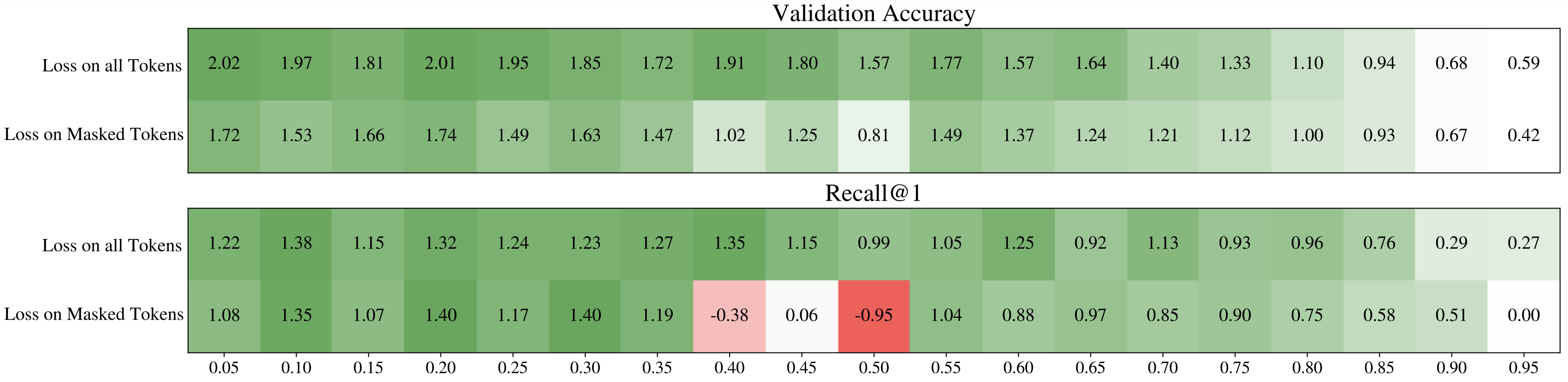}
\end{center}
\vspace{-5pt}
   \caption{The differences on ImageNet-1k validation accuracy (top table) and ImageNet KNN@1 (bottom table) relative to the baseline ViT-B/14 model for different masking ratio after training on ImageNet-1k for 300 epochs. The top row in both tables is when computing loss on all predicted tokens (the setup in the paper), whereas the bottom row is when only computing loss on the predicted masked tokens. Positive values, highlighted in different shades of green, indicate gains over the baseline performance. Red values denote the opposite. Baseline values are presented in Tables \ref{table:classification_eval} and \ref{table:knn_eval}.}
\label{fig:fig6}
\end{figure*}

In this section, we ablate apply MIM loss, defined in Equation \ref{eq2}, on only masked tokens. The comparison with our setup which applies MIM loss on all tokens is shown in \ref{fig:fig6}, where we study the effect of masking ratio when using a decoder with one transformer layer on a ViT-B encoder. We see that only predicting unmasked tokens results in suboptimal performance compared to our current setup in both ImageNet-1k classification and retrieval tasks.
\begin{table*}[t!]
\centering
\scalebox{0.54}{
\begin{tabular}{c|c|c|cccccccc|ccccc|ccccccccc}
\toprule
\multicolumn{2}{c|}{Model} & Training Setup & \rotatebox[origin=c]{90}{\textcolor{BrickRed}{Caltech 101}} & \rotatebox[origin=c]{90}{\textcolor{BrickRed}{CIFAR-100}} & \rotatebox[origin=c]{90}{\textcolor{BrickRed}{DTD}} & \rotatebox[origin=c]{90}{\textcolor{BrickRed}{Flowers102}} & \rotatebox[origin=c]{90}{\textcolor{BrickRed}{Pets}} & \rotatebox[origin=c]{90}{\textcolor{BrickRed}{Sun397}} & \rotatebox[origin=c]{90}{\textcolor{BrickRed}{SVHN}} &
\rotatebox[origin=c]{90}{\textcolor{BrickRed}{Mean}} &
\rotatebox[origin=c]{90}{\textcolor{ForestGreen}{Camelyon}} & \rotatebox[origin=c]{90}{\textcolor{ForestGreen}{EuroSAT}} & \rotatebox[origin=c]{90}{\textcolor{ForestGreen}{Resisc45}} & \rotatebox[origin=c]{90}{\textcolor{ForestGreen}{Retinopathy}} &
\rotatebox[origin=c]{90}{\textcolor{ForestGreen}{Mean}} &
\rotatebox[origin=c]{90}{\textcolor{RoyalBlue}{Clevr-Count}} & \rotatebox[origin=c]{90}{\textcolor{RoyalBlue}{Clevr-Dist}} & \rotatebox[origin=c]{90}{\textcolor{RoyalBlue}{DMLab}} & \rotatebox[origin=c]{90}{\textcolor{RoyalBlue}{dSpr-Loc}} & \rotatebox[origin=c]{90}{\textcolor{RoyalBlue}{dSpr-Ori}} & \rotatebox[origin=c]{90}{\textcolor{RoyalBlue}{KITTI-Dist}} & \rotatebox[origin=c]{90}{\textcolor{RoyalBlue}{sNORB-Azim}} & 
\rotatebox[origin=c]{90}{\textcolor{RoyalBlue}{sNORB-Elev}} &
\rotatebox[origin=c]{90}{\textcolor{RoyalBlue}{Mean}}\\
\midrule
\multirow{9}*{\rotatebox[origin=c]{90}{ViT-B/14}}& Baseline & \multirow{3}*{1N-1k$_{300}$} & 93.51 & 88.09 & 73.03 & 95.98 & 94.58  & 76.15 & 96.42 & 88.25 & 89.09 & 98.78 & 96.64 & 83.33 & 91.96 & 96.41 & 84.83 & 70.45 & 96.40 & 82.23 & 82.77 & 24.93 & 55.03 & 74.13\\
& Ours& & 93.85 & 88.71 & 72.87 & 96.86 & 94.22 & 76.28 & 96.28 & 88.44 & 86.78 & 98.80 & 96.21 & 83.56 & 91.34 & 95.84 & 82.80 & 73.70 & 97.91 & 83.58 & 84.46 & 24.65 & 57.66 & 75.08  \\
 & $\Delta$ & & \textbf{+0.34} & \textbf{+0.62} & -0.16 & \textbf{+0.87} & -0.36 & \textbf{+0.13} & -0.15 & \textbf{+0.19} & -2.31 & \textbf{+0.02} & -0.43 & \textbf{+0.23} & -0.62 & -0.57 & -2.03 & \textbf{+3.26} & \textbf{+1.51} & \textbf{+1.35} & \textbf{+1.70} & -0.28 & \textbf{+2.63}  & \textbf{+0.95} \\
 \cmidrule{2-25}
& Baseline & \multirow{3}*{1N-21k$_{30}$ FT} &  94.47 & 91.62 & 77.40 & 99.05 & 95.06 & 81.32 & 96.21 & 90.74 & 89.15 & 98.68 & 96.26 & 82.86 & 91.74 & 87.90 & 86.66 & 71.80 & 93.37 & 73.79 & 80.68 & 25.60 & 47.87 & 70.96 \\
& Ours & & 95.07 & 91.02 & 77.50 & 99.55 & 94.74 & 81.40 & 96.29 & 90.80 & 89.72 & 98.46 & 96.40 & 83.52 & 92.03 & 88.23 & 86.13 & 75.13 & 96.90 & 77.95 & 83.62 & 24.14 & 47.99 & 72.51 \\
 & $\Delta$ & &  \textbf{+0.60} & -0.60 & \textbf{+0.10} & \textbf{0.50} & -0.32 & \textbf{+0.08} & \textbf{+0.08}  & \textbf{+0.06} & \textbf{+0.57} & -0.22 & \textbf{+0.14} & \textbf{+0.66} & \textbf{+0.28} & \textbf{+0.33} & -0.53 & \textbf{+3.33} & \textbf{+3.53} & \textbf{+4.15} & \textbf{+2.94} & -1.46 & \textbf{+0.12} & \textbf{+1.55} \\
 \cmidrule{2-25}
& Baseline & \multirow{3}*{1N-21k$_{300}$ FT} & 94.35 & 91.92 & 76.92 & 99.37 & 94.82 & 82.36 & 96.58 & 90.90 & 88.29 & 98.96 & 96.73 & 83.20 & 91.80 & 94.01 & 86.51 & 73.98 & 98.11 & 82.91 & 81.64 & 24.64 & 51.66 & 74.18 \\
& Ours& & 94.58 & 91.59 & 77.77 & 99.51 & 95.12 & 82.61 & 96.76 & 91.13 & 89.95 & 98.56 & 96.57 & 83.72 & 92.20 & 95.25 & 85.52 & 75.62 & 99.87 & 83.38 & 84.32 & 26.20 & 50.08 & 75.03 \\
 & $\Delta$ & & \textbf{+0.23} & -0.33 & \textbf{+0.85} & \textbf{+0.14} & \textbf{+0.30} & \textbf{+0.25} & \textbf{+0.18} & \textbf{+0.23} & \textbf{+1.66} & -0.40 & -0.16 & \textbf{+0.52} & \textbf{+0.40} & \textbf{+1.24} & -0.99 & \textbf{1.64} & \textbf{+1.76} & \textbf{+0.47} & \textbf{+2.68} & \textbf{+1.56} & -1.58 & \textbf{+0.85} \\
\midrule
\multirow{6}*{\rotatebox[origin=c]{90}{ViT-L/14}} &Baseline &  \multirow{3}*{1N-21k$_{30}$ FT}& 94.77 & 92.27 & 78.1 & 99.22 & 93.97 & 81.72 & 95.91 & 90.85 & 90.01 & 98.72 & 96.5 & 82.06 & 91.82 & 87.78 & 84.22 & 72.26 & 93.57 & 82.85 & 83.52 & 24.97 & 47.33 & 72.06 \\
 &Ours & & 94.39 & 91.24 & 78.1 & 99.46 & 94.9 & 81.38 & 95.91 & 90.77 & 90.84 & 98.28 & 96.17 & 82.75 & 92.00 & 90.45 & 81.12 & 71.4 & 94.48 & 81.88 & 82.95 & 26.26 & 42.21 & 71.35 \\
 & $\Delta$ & & -0.38 & -1.03 & +0.00 & \textbf{+0.24} & \textbf{+0.93} & -0.34 & +0.00 & -0.08 & \textbf{+0.83} & -0.44 & -0.33 & \textbf{+0.69} & \textbf{+0.18} & \textbf{+2.67} & -3.10 & -0.86 & \textbf{+0.91} & -0.97 & -0.57 & \textbf{+1.29} & -5.12 & -0.71 \\
 \cmidrule{2-25}
 &Baseline &  \multirow{3}*{1N-21k$_{300}$ FT} & 94.13 & 93.33 & 77.67 & 99.43 & 94.41 & 80.98 & 96.83 & 90.97 & 89.38 & 99.00 & 97.12 & 82.01 & 91.88 & 89.75 & 83.38 & 70.23 & 97.27 & 85.60 & 82.81 & 27.24 & 54.27 & 73.82 \\
 &Ours & & 94.23 & 93.17 & 76.81 & 99.61 & 94.30 & 81.02 & 96.69 & 90.83 & 90.29 & 98.87 & 97.00 & 82.95 & 92.28 & 97.52 & 78.57 & 72.09 & 97.99 & 85.42 & 85.23 & 25.54 & 51.50 & 74.23 \\
 & $\Delta$ & & \textbf{+0.10} & -0.16 & -0.86 & \textbf{+0.18} & -0.11 & \textbf{+0.04} & \textbf{-0.14} & -0.14 & \textbf{+0.91} & \textbf{-0.13} & -0.12 & \textbf{+0.94} & \textbf{+0.40} & \textbf{+7.77} & -4.81 & \textbf{+1.86} & \textbf{+0.72} & -0.18 & \textbf{+2.42} & -1.70 & -2.77 & \textbf{+0.41} \\
\bottomrule
\end{tabular}
}
\vspace{-5pt}
\caption{Evaluation on all the 19 VTAB benchmarks under \textcolor{BrickRed}{natural}, \textcolor{ForestGreen}{specialized}, \textcolor{RoyalBlue}{structured} groups, following \cite{zhai2020vtab}. Values are in percentage.}
\label{table:vtab_eval_full}
\end{table*}

\end{document}